\documentclass[sigconf]{acmart}

\copyrightyear{2022} 
\acmYear{2022} 
\setcopyright{acmcopyright}\acmConference[MM '22]{Proceedings of the 30th ACM International Conference on Multimedia}{October 10--14, 2022}{Lisboa, Portugal}
\acmBooktitle{Proceedings of the 30th ACM International Conference on Multimedia (MM '22), October 10--14, 2022, Lisboa, Portugal}
\acmPrice{15.00}
\acmDOI{10.1145/3503161.3547912}
\acmISBN{978-1-4503-9203-7/22/10}

\usepackage{multirow}
\usepackage{booktabs} 
\begin{document}

\title{Weakly Supervised Video Salient Object Detection \\ 
via Point Supervision}



\author{Shuyong Gao\textsuperscript{\rm 1}, Haozhe Xing\textsuperscript{\rm 2} , Wei Zhang\textsuperscript{\rm 1,}*, Yan Wang\textsuperscript{\rm 2}, Qianyu Guo\textsuperscript{\rm 1}, Wenqiang Zhang\textsuperscript{\rm 1,2,3,}} 
\authornote{Corresponding authors}

\affiliation{%
 \institution{\textsuperscript{\rm 1}Shanghai Key Laboratory of Intelligent Information Processing, \\ School of Computer Science, Fudan University, Shanghai, China}
 \city{\textsuperscript{\rm 2}Academy for Engineering \& Technology, Fudan University, Shanghai, China \\ \textsuperscript{\rm 3}Yiwu Research Institute of Fudan University, Yiwu City, Zhejiang, China}
 \country{\{sygao18,hzxing21,weizh,yanwang19,qyguo20,wqzhang\}@fudan.edu.cn}
}

\renewcommand{\shortauthors}{Shuyong Gao et al.}



\begin{abstract}


Video salient object detection models trained on pixel-wise dense annotation have achieved excellent performance, yet obtaining pixel-by-pixel annotated datasets is laborious. Several works attempt to use scribble annotations to mitigate this problem, but point supervision as a more labor-saving annotation method (even the most labor-saving method among manual annotation methods for dense prediction), has not been explored. In this paper, we propose a strong baseline model based on point supervision.
To infer saliency maps with temporal information, we mine inter-frame complementary information from short-term and long-term perspectives, respectively. Specifically, we propose a hybrid token attention module, which mixes optical flow and image information from orthogonal directions, adaptively highlighting critical optical flow information (channel dimension) and critical token information (spatial dimension).
To exploit long-term cues, we develop the Long-term Cross-Frame Attention module (LCFA), which assists the current frame in inferring salient objects based on multi-frame tokens.
Furthermore, we label two point-supervised datasets, P-DAVIS and P-DAVSOD, by relabeling the DAVIS and the DAVSOD dataset.
Experiments on the six benchmark datasets illustrate our method outperforms the previous state-of-the-art weakly supervised methods and even is comparable with some fully supervised approaches.
Source code and datasets are available.
\end{abstract}





\begin{teaserfigure}
  \includegraphics[width=\textwidth]{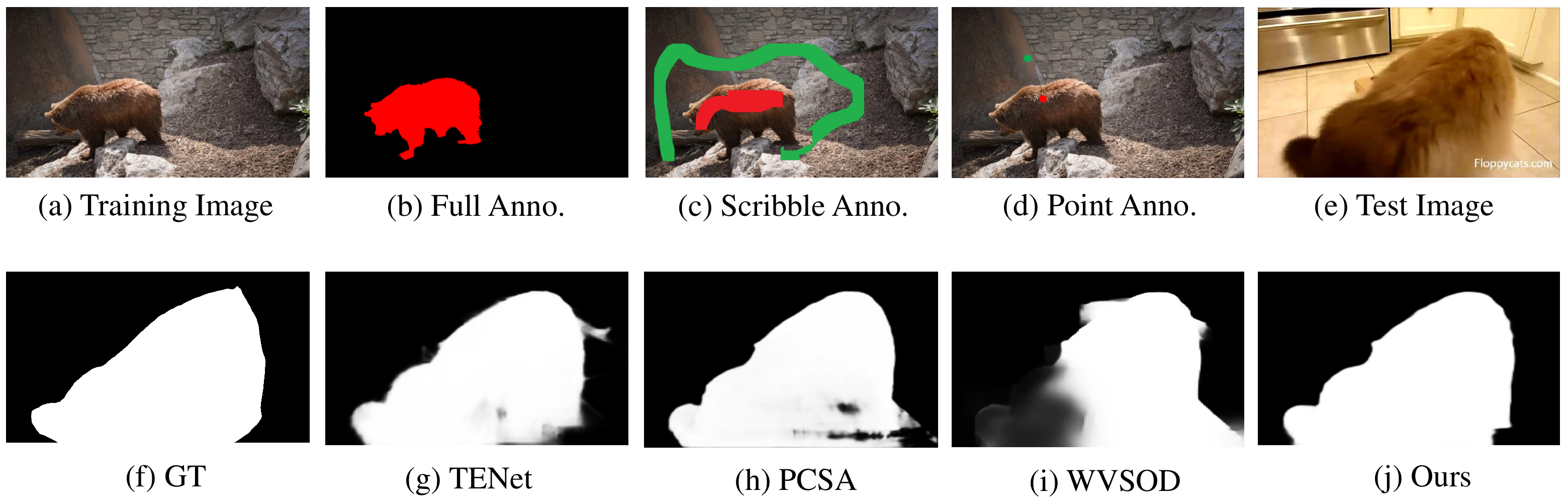}
  \caption{(a)-(d) show a visual comparison of the annotation methods. (e) denotes a frame of the video in the test dataset. (f)-(i) are the results from the typical full/scribble annotation based models, and (j) is our result.}
  \Description{xx.}
  \label{point_anno}
\end{teaserfigure}

\maketitle

\section{Introduction}

Video salient object detection (VSOD) aims to detect visually distinguishable regions of videos to mimic the human attention mechanism that pick out the important and required information from the complex external environment. VSOD has recently drawn broad attention in image processing and Computer Vision (CV) and is widely used in various CV tasks, such as video object segmentation \cite{full_vos, easyfram_vos}, video captioning \cite{videocaption}, and video compression \cite{videocompression1, videocompression2}.

Although still-image based Salient Object Detection (SOD) models have progressed significantly, detecting salient objects in videos remains a challenging task. As the usual process for training VSOD models, the still-image dataset (DUTS \cite{duts}) is used to pre-train the VSOD models, which is then fine-tuned on the video salient object datasets (i.e. DAVIS \cite{davis16} and DAVSOD \cite{davsod}). The main reason is that, despite a large amount of data in the video dataset, the scene and object of each clip's data lack diversity.
Training the VSOD model trained on DUTS is to improve its adaptability to various scenes to some degree, since it is expensive to label multi-scene video datasets pixel by pixel.
Previous weakly supervised salient object detection methods mainly focus on the still-image salient object detection \cite{psod, scwssod, wssa, mfnet, msw}. Both fully supervised and weakly supervised salient object detection methods have achieved large advances. However, detecting salient objects from videos is another story, as video salient object detection is heavily reliant on the motion information of objects between frames.

Recent work \cite{wsvsod} proposes a baseline model for weakly supervised video salient object detection using scribble annotations and presents two scribble-annotation based datasets (i.e., DAVIS-S and DAVSOD-S). 
Point supervision, as a near-least-effortless labeling method, has recently been applied to dense prediction tasks \cite{psod, what_point, metric_point}, where \cite{psod} proposes a point-supervised approach to detect salient objects in still-image salient object detection. In this paper, we attempt to investigate the application of point supervision in video salient object detection and propose a hybrid model based on transformer and convolution. Figure 1 illustrates the comparison between our approach and other typical approaches.


The transformer has recently made a great breakthrough in a range of image processing tasks \cite{vit, swin, pyramid_trs, token_to_token, visual_saliency}. It employs the self-attention layers to obtain a global view and can greatly alleviate the long-distance dependency problem. 
Current approaches for saliency detection (both still images and videos) are mainly based on Convolutional Neural Network (CNN) methods, and several models \cite{tritransnet, visualsaliency, psod} use transformer to detect the salient objects, but they all deal with still-image problems (RGB or RGBD saliency).
However, in VSOD tasks, it is more crucial to take into account the motion information of the objects. Some works \cite{vsod_pyramid, wsvsod, implicit_cod,dynamic} use 3D/2D convolution or ConvLSTM \cite{convlstm} to consider multi-frame motion information. Due to the intrinsic properties of convolution, these methods often lack sufficient global information, and to compensate for this drawback, some works \cite{vsod_pyramid, implicit_cod} leverage 3D Non-local \cite{nonlocal} methods to complement the global cues.

Unlike them, we exploit a transformer-based module to fully exploit the global attention property of the transformer, which obtains sufficient global spatial-temporal information.
Specifically, we mine the inter-frame supplementary information from both short-term and long-term perspectives. In the short-term aspect, we design a Hybrid Token Attention module (HTA) to determine the weights of fusion features from orthogonal directions: considering that optical flow information is often erroneous, we design an RGB Optical Flow Attention (ROFA) module for adaptive determination of fusion weights for channel dimensional features. 
On the other hand, to weaken the effect of tokens in non-salient regions on salient objects, we design a Token Attention module that determines the importance of each token based on the global features of each token.
Optical flow can only complement short-term motion features. To utilize long-term motion information, we develop a Long-term Cross-Frame Attention (LCFA) module that leverages features of salient objects in other frames to complement or find salient objects in the current frame.

Our main contributions can be summarized as follows:
\begin{itemize}
    \item We introduce the first point-based weakly supervised video salient object detection model, and build two point-supervised salient object detection datasets to validate our model, namely P-DAVIS and P-DAVSOD.
    
    \item 
    We propose a hybrid token attention module that extracts short-term motion information from two orthogonal directions, the channel dimension and the spatial dimension. To infer salient objects with long-term inter-frame information, we propose a Long-term Cross-Frame Attention module that optimizes the current frame token with multi-frame tokens. Extensive ablation experiments demonstrate the effectiveness of the proposed module.
    
    \item We conduct broad experiments on six wide-used benchmarks, and compare our method with 14 state-of-the-art full/weakly supervised salient object detection models under three evaluation metrics. Experiments demonstrate our model generally outperforms the models with stronger supervision and even outperforms fully supervised models in some metrics.
\end{itemize}




\section{Related Work}

\begin{figure*}[h]
	\centering
	\includegraphics[width=0.9\linewidth]{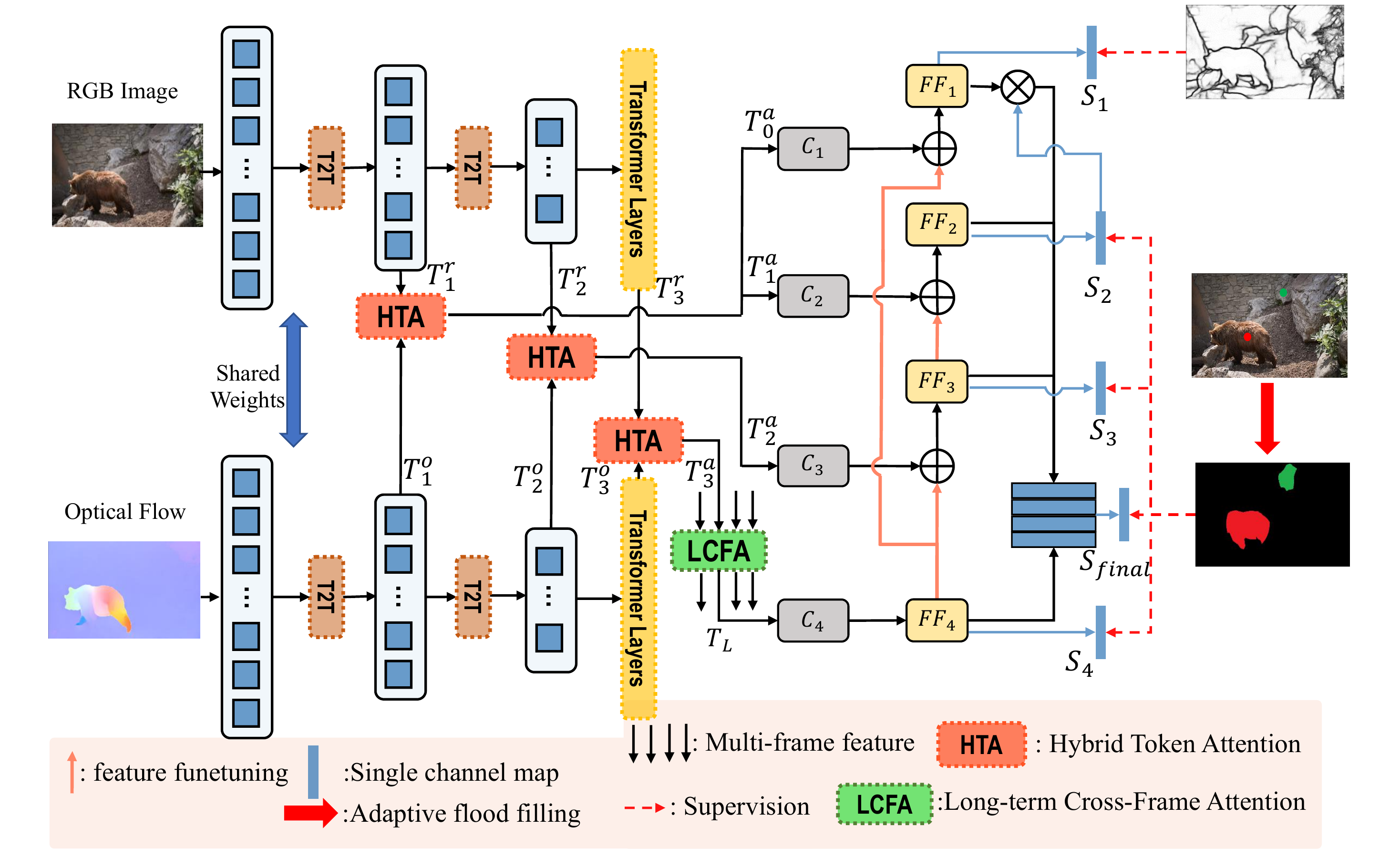}
	\caption{The framework of our proposed network. 
		T2T denotes the Token-to-Token  module.
	Black arrows indicate feature flow. We use the adaptive flood filling \cite{psod} to transform the point labels into the initial training labels. The upper and lower transformer backbone share weights to extract multi-scale image token features and optical flow token features respectively.}

	\label{pipline}
\end{figure*}

\subsection{Weakly-supervised Salient Object Detection}
To reduce the labeling cost, some methods propose weakly supervised or unsupervised methods to detect salient objects. For salient object detection in still-images, some approaches \cite{wsi,wssa} further optimized the results by incorporating dense Conditional Random Field (CRF) \cite{crf} as post-processing steps. Zeng et al. \cite{msw} use multimodal data, including category labels, captions, and noisy pseudo labels to train the saliency model where class activation maps (CAM) \cite{cam} is used as the initial saliency maps. Recently, Zhang et al. \cite{wssa} propose the scribble annotation based saliency detection model. Further, Yu et al. \cite{scwssod} leverage the gated CRFloss \cite{gate_crf} and saliency structure consistency loss to assist in training model using the scribble annotation. Gao et al. \cite{psod} proposes to train saliency models using point-supervised data and introduce a point-supervised dataset P-DUTS.
For detecting salient objects in video, some traditional VSOD methods detect salient objects based on the handcrafted feature in an unsupervised manner \cite{trad_vsod1, trad_vsod2}. Then the learning-based methods have achieved remarkable results. Tang et al. \cite{vsod_weakly} uses the existing saliency model to assist in generating pseudo-labels and combine hand-labeled data as training data. Yan \cite{vsod_semi} use optical flow to generate pseudo-labels with sparse frame annotations. Li et al. \cite{vsod_plug} retrain the pre-trained model by leveraging generated pseudo labels in the testing phase.
But high-quality labeled data is still inevitable to obtain the quality pseudo labels. Zhao et al. \cite{wsvsod} introduces the scribble annotation datasets and uses them for fully weakly supervised training, which greatly reduces annotation time and achieves excellent performance.

Unlike the above approaches, we introduce the point-supervised datasets and employ them to train the model. To the best of our knowledge, this is the first work to detect video salient objects based on point supervision.

\subsection{Point Supervision}
Point supervision has been studied in recent years as an extremely cost-saving labeling method, but it poses significant hurdles to dense labeling tasks due to 
its extremely sparse properties. Gao et al. \cite{psod} propose a point-supervised still-image salient object detection model and propose an adaptive flood filling algorithm to generate initial pseudo-labels. There has been some researches on point annotations in weakly supervised segmentation \cite{what_point, metric_point} and instance segmentation \cite{point_large, point_latent, point_extreme, point_regional}. Bearman et al. \cite{what_point} introduce point annotation by asking annotators to point to an object and incorporate the labeled point with an objectness potential in the loss function. Point-based instance segmentation is usually employed in an interactive manner where the models are trained with full supervision and Interactively use user annotation points during the test phase.
In addition, point supervision is often used in tasks such as object localization \cite{object_loca1, object_loca2}, crowd counting \cite{crowd_count_nwpu, object_loca2} and pose estimation \cite{pose_est_deep, pose_est_ijcv}. Most of them do not require pixel-by-pixel prediction.

\subsection{Transformer in Vision Tasks}

The transformer \cite{attent_is_all} from Natural Language Processing (NLP) has recently gained a lot of attention in visual tasks. Some works \cite{deformable_detr, detr, panoptic_tr} employ convolution layers to extract feature maps and then use the transformer layer to acquire global information. A large number of works employ the transformer structure to extract semantic features directly. Vision Transformer (ViT) \cite{vit} introduces a pure transformer architecture in image classification tasks and achieved outstanding results. Considering the scale difference between natural language and images, Liu et al. \cite{swin} develop a shifted window scheme that greatly improves computational efficiency. Wang et al. \cite{pyramid_trs} propose a pyramid transformer that enables the transformer structure to adapt to a variety of dense prediction vision tasks. 
\cite{tritransnet,visual_saliency,psod} employ the transformer as the backbone, and use convolution or transformer layers for decoding to predict salient maps in RGB-D or RGB images.
Yuan et al. \cite{token_to_token} aggregates adjacent tokens to facilitate modeling local features by introducing a token-to-token module to transformer layers and is capable of producing multi-scale features. 
In this paper, T2T-ViT \cite{token_to_token} is used to extract multi-scale features of every single frame considering its high efficiency and ability to extract multi-scale features. And closely following the token features extracted by T2T-ViT, we build the hybrid token attention module to fuse short-term motion features and the long-term cross-frame attention module to aggregate multi-frame token features.

\section{Methodology}

\subsection{Transformer Encoder}
In this paper, "T2T-ViT\_t-14" (21.5M parameters) is used as our backbone, as it has a similar number of parameters to the frequently used ResNet-50 \cite{resnet} and VGG-16 \cite{vgg} backbone in salient object detection and they are both pre-trained on ImageNet \cite{imagenet}. T2T transformer is composed of cascaded Token-to-Token (T2T) modules, and each T2T module consists of two steps, i.e., Re-structurization and Soft Split.


\noindent \textbf{Re-structurization.} As shown in Fig \ref{pipline}, given a token sequence $L$, it will perform the following self-attention transformation:
\begin{equation} \label{xx}
L^{'}=MLP(MSA(L))
\end{equation}

\noindent where $MSA(\cdot)$ refers to multihead self-attention operation and $MLP(\cdot)$ refers to  the multilayer perceptron in standard Transformer \cite{vit}. Then the obtained $L^{'} \in \mathbb{R}^{l \times c}$ will be reshaped as a 3D tensor $L^{''} \in \mathbb{R}^{h \times w \times c}$: $L^{''}=Re(L^{'})$, where $Re(\cdot)$ denotes reshape operation,  $h,w,c$ are height, width, and the number of channels, respectively, and $l=h \times w$.

\noindent \textbf{Soft Split.}
After obtaining $L^{''}$, the soft split operation $SS(\cdot)$ is employed to build local information and decrease the token length. $L^{''}$ is split into patches with overlapping, which introduce a convolution-like local prior. By applying soft split, each new patch shares features with surrounding patches and is, therefore, more correlated. The tokens in each patch are unfold as one token, and the output tokens are fed into the next Token-to-Token module.

\noindent \textbf{T2T module.} Re-structurization and soft split are applied iteratively, which can be formulated as follow:
\begin{gather} \label{xx}
\begin{split}
&L_i^{'} = MLP(MSA(L_i))\\
&L_i^{''} = Re(L_i^{'})\\
&L_{i+1} = SS(L_i^{''}) 
\end{split}
\end{gather}

Given the input image $L_0$, a soft split is first adopted: $L_1$ = SS($L_0$). $L_1$ is then fed into two cascaded T2T modules, the tokens output by each T2T module can be denoted as $L_2$ and $L_3$. Finally, $L_3$ is added with position embedding, and the result is fed into fourteen transformer layers, which output its final result $L_4$ of T2T-ViT. $L_2, L_3$ and $L_4$ are used by subsequent parts of the network as shown in Fig. \ref{pipline}, which are re-denoted as $T_1^r, T_2^r, T_3^r$ (or $T_1^o, T_2^o, T_3^o$), respectively. In this paper, the T2T-ViT with shared weights is used to encode RGB images and optical flow. The optical flow maps are obtained by leveraging the widely-used optical flow estimation network, FlowNet2.0 \cite{flownet2}.


\subsection{Hybrid Token Attention}


\subsubsection{RGB Optical Flow Attention}

We explicitly incorporate optical flow to leverage the short-term inter-frame features. Nevertheless, optical flow information is frequently erroneous, and a more appropriate approach is to identify the effective portion of the optical flow and RGB image information dynamically. As shown on the left side of Fig. \ref{fusat}, we develop an RGB Optical Flow Attention (ROFA) module to explicitly assess the importance of each channel in image token and flow token, which are detailed as follows.
\begin{gather} \label{xx}
\begin{split}
&T_i = Cat(T_i^r, T_i^o)\\
&y_i^1 = \sigma(W_i^{f_1} \times LN(T_i))\\
&y_i^2 = Sig(W_i^{f_2} \times y_i^1)\\
&T_i^f = LN(T_i) + y_i^2 \odot LN(T_i),\\
\end{split}
\end{gather}

\noindent where $i$ denotes the $i$ th stage ($i=1,2,3$), $T_i^r$ and $T_i^o$ denote the token of the RGB image and optical flow, respectively. $Cat(T_i^r, T_i^o)$ means concatenating $T_i^r$ and $T_i^o$ along the channel direction, $LN(\cdot)$ denotes layer normalization, $\sigma(\cdot)$ denotes GeLU function, $Sig(\cdot)$ denotes sigmoid activation. $W_i^{f_1}$ and $W_i^{f_2}$ denote the weights of two cascaded fully connected layers. The symbols $\times$ and $\odot$ represent the matrix multiplication and element-wise multiplication respectively.

It can be seen that $y_i^2$ aggregates the information of the image token and optical flow token. For any pair of image token and optical flow token combinations, $y_i^2$ outputs the weights of each position of $T_i$ according to its combination. Unimportant features can be filtered out by $y_i^2 \times LN(T_i)$, and a skip connection is applied to facilitate optimization.

\subsubsection{Token Attention}

\begin{figure*}[h]
	\centering
	\includegraphics[width=0.90\linewidth]{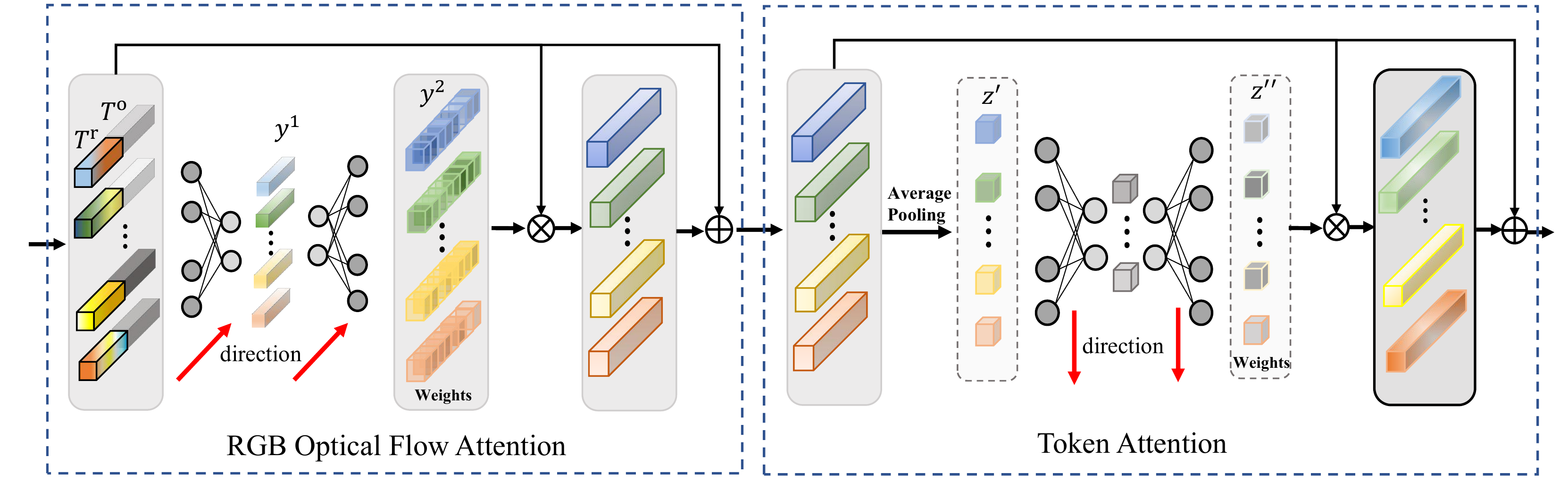}
	\caption{Hybrid Token Attention (HTA) consists of RGB Optical Flow Attention (ROFA) module and Token Attention (TA) module.}
	\label{fusat}
\end{figure*}

RGB Optical Flow Attention considers the importance of the fused features from the channel dimension. Token Attention (TA) considers the importance of each token from the spatial dimension of the image (right of Fig. \ref{fusat}). It is a token-oriented SE module \cite{senet} with a residual connection that outputs the importance of each token based on the global feature $z_i^{'}$, explicitly emphasizing or suppressing each token with $z_i^{''}$. Specifically, given the token features $T_i^f$ output by ROFA, first, a 1D global average pooling is performed to obtain the representative features $z_i^{'}$ of each one-dimensional token. Then two cascaded fully connected layers are used, where the first is followed by the GeLU activation and the second is followed by the sigmoid activation. It can be expressed as follows.
\begin{gather}
\begin{split}
    &z_i^{'} = Avg(T_i^f)\\
    &z_i^{''} = Sig(W_i^{a_2} \times \sigma(W_i^{a_1} \times z_i^{'}))\\
    &T_i^a = z_i^{''} \odot LN(T_i^f) + LN(T_i^f),
\end{split}
\end{gather}
\noindent where $Avg(\cdot)$ denotes global average pooling. $W_i^{a_1}$ and $W_i^{a_2}$ denote the weights of the two cascaded fully connected layers. $z_i^{''}$ denotes the importance of each token feature that mixes RGB token and optical flow token, which varies as per different $T_i^f$. The important token is highlighted explicitly by $z_i^{''} \odot LN(T_i^f)$, and finally a constant connection is added to prevent excessive information loss. Note that $T_1^r$ and $T_1^o$ are fed into two separate HTA modules which produce $T_0^a$ and $T_1^a$ respectively.

\subsection{Long-term Cross-Frame Attention}

\begin{figure}[h]
	\centering
	\includegraphics[width=0.65\linewidth]{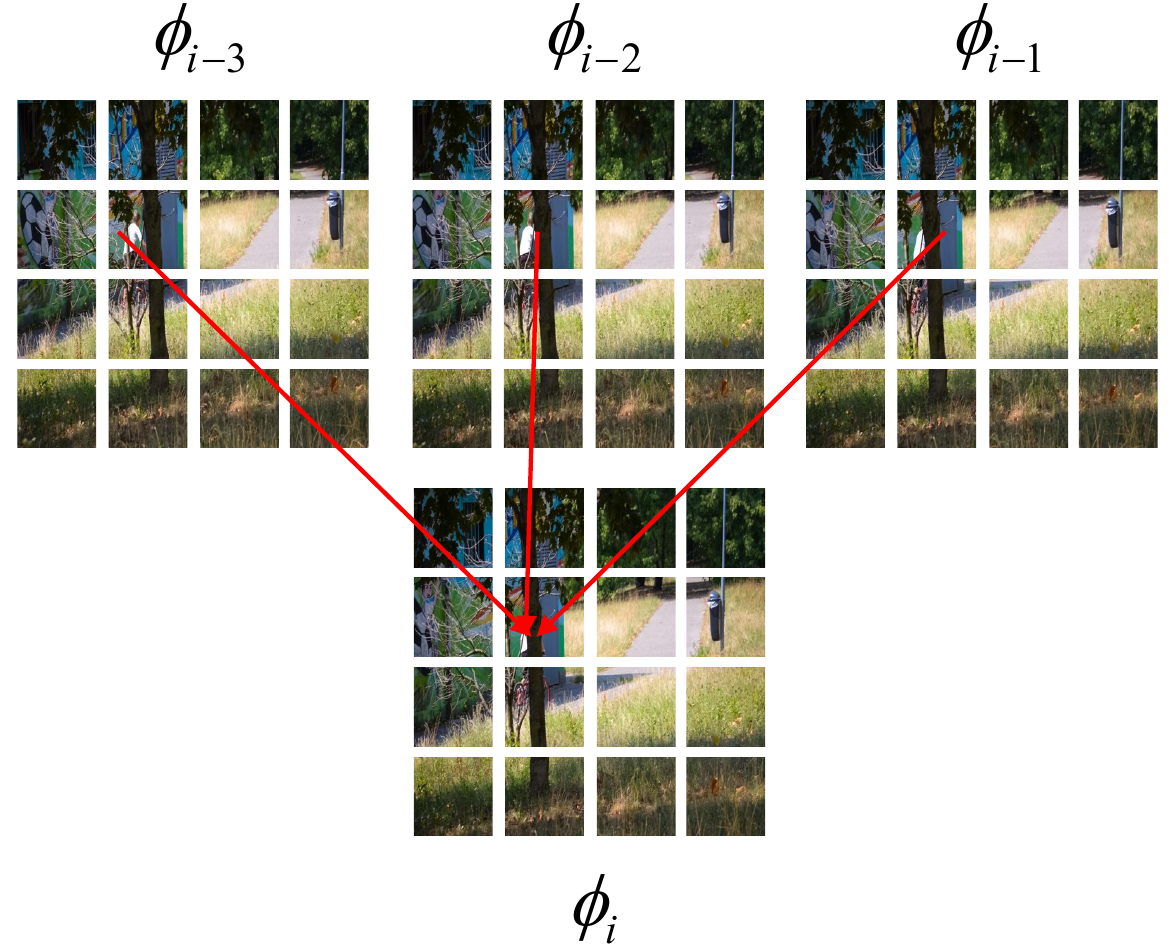}
	\caption{Illustration of the Long-term Cross-frame Attention module. The patches of $\phi_{i-3}$, $\phi_{i-2}$, $\phi_{i-1}$, $\phi_{i}$ represent multi-frame token features. The tokens in current frame $\phi_{i}$ can leverage tokens from other frames to assist $\phi_{i}$ in detecting salient objects.}
	\label{lcfa_visual}
\end{figure}

Compared with the salient objects in still-images, salient objects in videos are frequently blurred and obscured, and even people can only infer salient objects by comparing multiple frames. 
By employing ROFA, the optical flow is incorporated as the short-time motion feature of adjacent frames. To infer salient items using long-term inter-frame information, we propose a Long-term Cross-Frame Attention module (LCFA).
Long-range dependence within images has been explored extensively. In video salient object detection, long-range dependence between multiple frames still exists. 
For example, as shown in Fig. \ref{lcfa_visual}, due to occlusion, objects that were salient in the earlier frames are difficult to find in the current frame. At this point, multi-frame-based similarity must be utilized, since the objects are no longer salient in the current frame.

Specifically, LCFA takes the token features $T_3^a$ of $n$ consecutive frames. For simplicity, we use $\phi_i$ to denote the $T_3^a$ of the i-th frame, so the input of LCFA can be denoted as $\{\phi_i \in \mathbb{R}^{l \times d} \lvert  i=1,2, \cdots, n \}$.
For the current frame feature $\phi_c$, we take all other frames except the current frame $\phi_o \in \mathbb{R}^{(n-1)l \times d}$ to reconstruct the current frame, which can employ other frames to boost the detection of the salient object in the current frame. To be specific, $\phi_c$ is embedded to query $Q_c \in \mathbb{R}^{l \times d}$ by employing one linear projection. A similar operation is applied to $\phi_o$ to acquire queries keys $K_o$, and values $V_o$. Then cross-frame attention is performed and a constant mapping is added to retain the features based on the self-attention of the current frame. 
\begin{gather} \label{xx}
\begin{split}
&\phi_c^{'}= \phi_c + softmax(\frac{Q_{c} (K_o)^T}{\sqrt{c}} )V_o\\
&\phi_c^{''} = LN(MLP(\phi_c^{'}))
\end{split}
\end{gather} 

\noindent where $MLP$ denotes two cascades fully connected layer followed by GeLU activation, and $LN$ denotes the layer norm operation. The cross-frame attention is implemented in the form of multi-head attention \cite{vit}, and the LCFA module is executed n times (n=3 in this paper). As shown in Fig. \ref{pipline}, LCFA takes in $T_3^a$ and outputs $\phi_c^{''}$ which we denote as $T_L$.

\subsection{Edge Complementary Decoder}
As shown in Fig. \ref{pipline}, we design a CNN-based decoder. It consists of a salient region detection part and a edge detection part. $T_0^a$, $T_1^a$,$T_2^a$ and $T_L$ are reshaped into 3D tensors and transformed into $C_1, C_2, C_3, C_4$ by four $1 \times 1$ convolution layers for channel transformation. All features are transformed into 64 channels. 
$C_1$ and $C_4$ are used to generate edge features to complement details. $C_2$, $C_3$ and $C_4$ are used to progressively decode the saliency maps. In Fig. \ref{pipline}, $FF=\{FF_{i}|i=1,2,3,4\}$ denotes the fused feature from two scales, which are employed to model the final saliency map $S_{final}$. Detailed steps are shown below.

\begin{equation}
FF_i=\left\{
	\begin{aligned}
	&ff_i(C_i+f_i(interp(FF_{4}))) \quad i=1\\
	&ff_i(C_i+f_i(interp(FF_{i+1}))) \quad i=2,3\\
	&ff_i(C_i) \quad\quad\quad\quad\quad\quad\quad\quad\quad i=4\\
	\end{aligned}
	\right.
\end{equation}

\noindent where $ff_i$ denotes three cascaded convolution layers, $f_i$ denotes two convolution layers for finetuning fused feature from last layer, $interp(\cdot)$ denotes bilinear interpolation. As shown in Fig. \ref{pipline}, $FF_i (i=1,2,3,4)$ is transformed into single channel maps $S_i (i=1,2,3,4)$, in which $S_i (i=2,3,4)$ are constrained by pseudo labels and $S_1$ is constrained by edge map. Note that we don't directly fuse these edge features, but leverage the Saliency Region as Guidance (SRG) to explicitly filter out the background region edges by $E = FF_1 \odot S_2$. Finally, $E ,F_2,F_3$ and $F_4$ are combined and transformed into the final saliency map $S_{final}$ by two cascaded convolution layers.








\subsection{Loss Function}
During the network training, we employ four loss functions: binary cross entropy loss, partial cross entropy loss \cite{partial_ce}, and gated CRF loss \cite{scwssod,gate_crf}, and smoothness loss \cite{scwssod}. For the edge detection part, we use binary cross entropy loss to constrain the output. Similar to the previous SOD models \cite{wssa, wsvsod}, we use the edge detector \cite{edge_richer} to generate the edges of the training images.
\begin{equation} \label{partial_ce}
\mathcal{L}_{bce}=-\sum_{r,c}[y_{r,c}log(\varepsilon_{r,c})+(1-y_{r,c})log(1-\varepsilon_{r,c})],
\end{equation}


\noindent where $y$ refers to the ground-truth edge map, $\varepsilon$ denotes the predicted edge map, $r$ and $c$ represent the row and column coordinates, respectively.

For the predicted salient object, we leverage partial cross entropy loss and smoothness loss to constrain $S_{final}$ like \cite{wsvsod}, and an additional gated CRF loss is used for local consistency. Partial cross entropy loss can be expressed as follows.
\begin{equation} \label{partial_ce}
\mathcal{L}_{pbce}=-\sum_{j\in J}[g_jlog(s_j)+(1-g_j)log(1-s_j)],
\end{equation}     
	
\noindent where $J$ represents the labeled area, $g$ refers to the ground truth, $s$ represents the predicted saliency map. The smoothness loss is defined as:
\begin{equation} \label{smooth_loss}
\mathcal{L}_{smooth} = \sum_{u,v}\sum_{d \in \vec{x},\vec{y}}|\partial_{d}S_{u,v}|e^{-|\partial_{d}I_{u,v}|} ,
\end{equation}

\noindent where $S_{u,v}$ is the salient value at $(u,v)$, $|\partial_{d}(.)|$ denotes the absolute value of directional derivative along $\vec{x}$ or $\vec{y}$. To obtain better object structure and edges, following \cite{scwssod}, gated CRF is used in our loss function:
\begin{equation} \label{xx}
\mathcal{L}_{gcrf} =  \sum_{i}\sum_{j\in K_i}  |s_i - s_j|f(i,j),
\end{equation}

\noindent where $K_i$ denotes the areas covered by $k \times k$ kernel around pixel $i$. 

\noindent  $s_i$ and $s_j$ are the saliency values of $s$ at position i and j, respectively.
 $f(i,j)$ refers to the Gaussian kernel bandwidth filter:
\begin{equation} \label{xx}
f(i,j)= \frac{1}{w}\exp(-\frac{\parallel PT(i)-PT(j) \parallel_{2}}{2\sigma_{PT}^2}-\frac{\parallel I(i)-I(j) \parallel_{2}}{2\sigma_I^2}),
\end{equation}

\noindent where $\frac{1}{w}$ is the weight for normalization, $I(\cdot)$ and $PT(\cdot)$ are the RGB value and position of pixel, respectively.  $\sigma_{PT}$ and  $\sigma_I$\ are the hyper parameters to control the scale of Gaussian kernels. 

During the training process, $S_1$ is constrained by $\frac{1}{5}\mathcal{L}_{bce}$, $S_2$, $S_3$ and $S_4$ is constrained by $\frac{1}{3}$$\mathcal{L}_{pbce}+\frac{1}{3}$$\mathcal{L}_{smooth}$, $S_{final}$ is constrained by $\mathcal{L}_{pbce}+0.3\mathcal{L}_{smooth}+0.1\mathcal{L}_{gcrf}$. The sum of all the loss functions is used as the final loss function.




\section{Experimental Results}

\subsection{Point-superviesd Video Dataset}
To verify the effectiveness of our algorithm, we re-label two widely used video salient object detection datasets, DAVIS \cite{davis16} and DAVSOD \cite{davsod}, using point annotation, and name them P-DAVIS and P-DAVSOD. As shown in Fig. \ref{point_anno}, for each salient object we only annotate one pixel location, and randomly select a point from the background as the background annotation point. Note that we don't use thick points, since thick points are inconvenient to label slender objects (In Fig. \ref{point_anno} we zoom in on the point annotation for clarity.). Six participants were invited to participate in the labeling task, and we randomly selected one of the six labels to reduce the individual bias.

\begin{table*}[htbp]
	\centering
	
	\resizebox{\textwidth}{!}{
		\begin{tabular}{cc |cccccccc ccc|cccc}
			\hline
			
			\multirow{3}{*}{Metric} & &\multicolumn{11}{|c|}{Fully Sup.Methods} &\multicolumn{4}{c}{Weakly Sup./Unsup. Methods} \\
			& &EGNet &PoolNet &MBNM &PDB &FGRN &MGA &RCRNet &SSAV &PCSA &TENet &DCFNet &GF &WSSA &WSVOD &Ours \\
			& &\cite{egnet} &\cite{poolnet} &\cite{mbnm} &\cite{pdb} &\cite{fgrn} &\cite{mga} &\cite{rcrnet} &\cite{ssav} &\cite{vsod_pyramid} &\cite{tenet} &\cite{dcfnet} &\cite{trad_vsod2} &\cite{wssa} &\cite{wsvsod}  \\
			\hline
			\multirow{4}{*}{SegV2} &$S_m$\hfill$\uparrow$             &0.845 &0.782 &0.809 &0.864 &-     &0.880 &0.843 &0.849      &0.866  &0.868 &0.893     &0.699  &0.733 &\textcolor{blue}{0.804}  &\textcolor{red}{0.834} (+3.0\%)\\
			&$F_\beta$\hfill$\uparrow$                          &0.774 &0.704 &0.716 &0.808 &-     &0.829 &0.782 &0.797      &0.811  &0.810 &0.837     &0.592  &0.664 &\textcolor{blue}{0.738}  &\textcolor{red}{0.753} (+1.5\%)\\
			&MAE\hfill$\downarrow$                                    &0.024 &0.025 &0.026 &0.024 &-     &0.027 &0.035 &0.023      &0.024  &0.025 &0.014     &0.091  &0.039 &\textcolor{blue}{0.033}  &\textcolor{red}{0.030} (-0.3\%)\\
			
			\hline
			\multirow{4}{*}{VOS} &$S_m$ \hfill$\uparrow$              &0.793 &0.773 &0.742 &0.818 &0.715 &0.791 &0.873 &0.786      &0.828  &0.845 &-     &0.615  &0.682 &\textcolor{blue}{0.750} &\textcolor{red}{0.796} (+4.6\%) \\
			&$F_\beta$ \hfill$\uparrow$                         &0.698 &0.709 &0.670 &0.742 &0.669 &0.734 &0.833 &0.704      &0.747  &0.781 &-     &0.506  &0.648 &\textcolor{blue}{0.666}  &\textcolor{red}{0.739} (+7.3\%) \\
			&MAE \hfill$\downarrow$                                   &0.082 &0.082 &0.099 &0.078 &0.097 &0.075 &0.051 &0.091      &0.065  &0.052 &-     &0.162  &0.106 &\textcolor{blue}{0.091} &\textcolor{red}{0.074} (-1.7\%)\\
			
			\hline
			\multirow{4}{*}{DAVIS} &$S_m$\hfill$\uparrow$             &0.829 &0.854 &0.887 &0.882 &0.838 &0.910 &0.886 &0.892      &0.902  &0.905 &0.914     &0.688  &0.795 &\textcolor{red}{0.828}  &\textcolor{blue}{0.808} (-2.0\%) \\
			&$F_\beta$\hfill$\uparrow$                          &0.768 &0.815 &0.861 &0.855 &0.783 &0.892 &0.848 &0.860      &0.880  &0.881 &0.900     &0.569  &0.734 &\textcolor{red}{0.779}  &\textcolor{blue}{0.754} (-2.5\%)\\
			&MAE\hfill$\downarrow$                                    &0.057 &0.038 &0.031 &0.028 &0.043 &0.023 &0.027 &0.028      &0.022  &0.017 &0.016     &0.100  &0.044 &\textcolor{red}{0.037}  &\textcolor{blue}{0.040} (+0.3\%)\\
			
			\hline
			\multirow{4}{*}{DAVSOD} &$S_m$\hfill$\uparrow$            &0.719 &0.702 &0.637 &0.698 &0.693 &0.741 &0.741 &0.755      &0.741  &0.779 &0.755     &0.553  &0.672 &\textcolor{blue}{0.705}  &\textcolor{red}{0.718} (+1.3\%)\\
			&$F_\beta$\hfill$\uparrow$                          &0.604 &0.592 &0.520 &0.572 &0.573 &0.643 &0.654 &0.659      &0.656  &0.697 &0.660     &0.334  &0.556 &\textcolor{blue}{0.605}  &\textcolor{red}{0.622} (+1.7\%)\\
			&MAE\hfill$\downarrow$                                    &0.101 &0.089 &0.159 &0.116 &0.098 &0.083 &0.087 &0.084	   &0.086  &0.070 &0.074     &0.167  &0.101 &\textcolor{blue}{0.103}  &\textcolor{red}{0.094} (-0.9\%)\\
			
			\hline
			\multirow{4}{*}{FBMS} &$S_m$\hfill$\uparrow$              &0.878 &0.839 &0.857 &0.851 &0.809 &0.908 &0.872 &0.879      &0.868  &0.916 &-     &0.651  &0.747 &\textcolor{blue}{0.778}  &\textcolor{red}{0.812} (+3.4\%)\\
			&$F_\beta$\hfill$\uparrow$                          &0.848 &0.830 &0.816 &0.821 &0.767 &0.903 &0.859 &0.865      &0.837  &0.915 &-    &0.571  &0.727 &\textcolor{blue}{0.786}  &\textcolor{red}{0.794} (+0.8\%)\\
			&MAE\hfill$\downarrow$                                    &0.044 &0.060 &0.047 &0.064 &0.088 &0.027 &0.053 &0.040      &0.040  &0.024 &-     &0.160  &0.083 &\textcolor{blue}{0.072}  &\textcolor{red}{0.055} (-1.7\%)\\

			\hline
			\multirow{4}{*}{ViSal} &$S_m$\hfill$\uparrow$             &0.946 &0.902 &0.898 &0.907 &0.861 &0.940 &0.922 &0.942      &0.946  &0.949 &0.952     &0.757  &0.853 &\textcolor{blue}{0.857}  &\textcolor{red}{0.878} (+2.1\%)\\
			&$F_\beta$\hfill$\uparrow$                          &0.941 &0.891 &0.883 &0.888 &0.848 &0.936 &0.907 &0.938      &0.941  &0.949 &0.953     &0.683  &0.831 &\textcolor{blue}{0.831}  &\textcolor{red}{0.858} (+2.7\%)\\
			&MAE\hfill$\downarrow$                                    &0.015 &0.025 &0.020 &0.032 &0.045 &0.017 &0.026 &0.021      &0.017  &0.012 &0.010     &0.107  &0.038 &\textcolor{blue}{0.041}  &\textcolor{red}{0.028} (+1.3\%)\\

			\hline
	\end{tabular}}
	\caption{Quantitative comparisons of $S_m$, $F_\beta$ and $MAE$ on six widely-used VSOD datasets. The top two results in weakly sup./unsup. methods are marked in red and blue font respectively. The content in parentheses indicate the performance gains.}
	\label{comparison_SOTA}
\end{table*}

\begin{figure*}[h]
	\centering
	\includegraphics[width=0.95\linewidth]{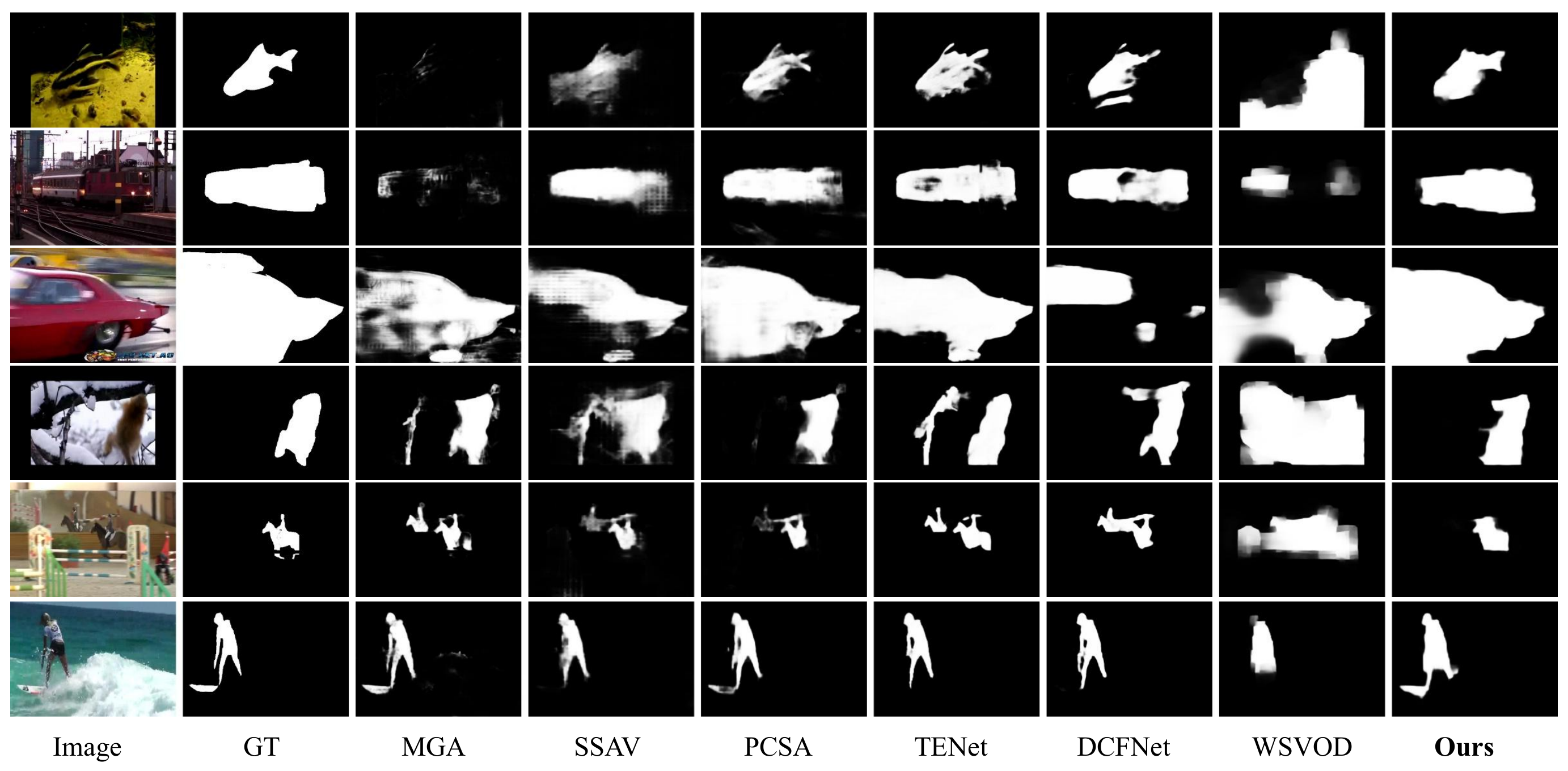}
	\caption{Qualitative comparison with different methods. Columns 3 to 7 are fully supervised methods, and the last two columns are weakly supervised methods.}
	\label{vision_compare}
\end{figure*}

\subsection{Experimental Setup}

\textbf{Datasets.} Similar to previous VSOD learning strategies, we pretrain our model on the point-based image SOD dataset \cite{psod} and our labeled point-based VSOD datasets, P-DAVIS and P-DAVSOD. Then we fine-tune model on P-DAVIS and P-DAVSOD. We conduct the experiments on six widely-used VSOD datasets, including DAVIS \cite{davis16}, DAVSOD \cite{davsod}, FBMS \cite{fbms}, ViSal \cite{visal}, SegTrack-V2 \cite{segtrack}, and VOS \cite{vos_dataset} to evaluate our model's performance.

\noindent \textbf{Evaluation Metrics.} In our experiments, we take three metrics to evaluate our model: mean absolute error ($MAE$), maximum F-measure ($F_{\beta}$), and structure-based metric ($S_{m}$) \cite{s-measure}. F-measure takes both precision and recall into account to comprehensively combine the precision and recall, which is defined as:
\begin{equation}
F_\beta=\frac{\left(1+\beta^2\right)\times Precision \times Recall}{\beta^2 \times Precision+Recall}
\end{equation}
\noindent where $\beta^2$ is commonly set to 0.3 to weight precision more than recall \cite{pr0.3}. $S_{m}$ focuses on evaluating the structural information of non-binary saliency map and the ground truth, which is defined as the weighted sum of object-aware $S_o$  and region-aware $S_r$ \cite{s-measure}. It can be computed as: $S_m=\gamma S_o+(1-\gamma)S_r$, where $\gamma$ is usually set to 0.5.

\subsection{Implementation Details}
The proposed model is implemented on the Pytorch toolbox. We train our model on a PC with four NVIDIA TITAN XP GPU (with 12GB memory). As shown in Fig. \ref{pipline}, our model takes as input RGB image and flow. Similar to \cite{wsvsod}, we employ a widely-used optical flow estimation model \cite{flownet2} to calculate the optical flow map, and for static images (P-DUTS), we make two identical images input \cite{flownet2} to obtain the optical flow image. Adaptive flood filling \cite{psod} is employed to generate pseudo-labels for training, where the hyperparameter $\gamma$ (the radius of the circle) is set to 1/6.
Adam optimizer and polynomial learning strategy are used both in the pre-training and finetune phase. The input images are resized to $224 \times 224$. Horizontal flip and random crop are used as data augmentation. It takes 20 epochs and 10 epochs for the pre-train and finetune procedures respectively. During the pre-training phase, LCFA was not assembled. The learning rate is set to $1 \times 10^{-4}$. The batch size is set to 48. In the finetune phase, LCFA is trained jointly with other modules, whose learning rate is set to $1 \times 10^{-4}$ and other parameters are set to $1 \times 10^{-5}$. The batch size is set to eight, and the length of the frames per batch is set to four. During testing, we resized each image to 224 $\times$ 224 and then feed it to the model to predict the final saliency maps without any post-processing.


\subsection{Comparisons with the State-of-the-arts} 

\begin{table*}
\centering
\label{tab:freq}
\scalebox{0.85}{
	\begin{tabular}{l|ccc| ccc| ccc}
		\hline
		\multirow{2}{*}{Model}    &\multicolumn{3}{c|}{DAVIS}   &\multicolumn{3}{c|}{SegV2} &\multicolumn{3}{c}{DAVSOD}\cr
		&MAE$\downarrow$  &$F_\beta \uparrow$ &$S_m \uparrow$  &MAE$\downarrow$  &$F_\beta \uparrow$ &$S_m \uparrow$ &MAE$\downarrow$  &$F_\beta \uparrow$ &$S_m \uparrow$\\
		\hline
		(a) B             &0.061 &0.753	&0.758 &0.044 &0.769 &0.78 &0.117 &0.624 &0.688\\
		(b) B+SRG         &0.057 &0.737 &0.768  &0.043 &0.736 &0.8 &0.124 & 0.606 &0.678\\
		\hline
		(c) B+SRG+ROFA     &0.058 &0.734 &0.769 &0.043 &0.731 &0.784 &0.112 &0.614 &0.691 \\
		(d) B+SRG+TA      &0.054 &0.753 &0.778 &0.042 &0.748 &0.8 &0.111 &\textcolor{red}{0.63} &0.696 \\
		(e) B+SRG+HTA (ROFA+TA) &0.052 &\textcolor{red}{0.762} &0.783 &0.04 &\textcolor{red}{0.758} &0.798 &0.11 &\textcolor{blue}{0.628} &0.699 \\                              
		\hline
		(f) B+SRG+HTA+ConvLSTM   &\textcolor{blue}{0.042} &\textcolor{blue}{0.754} &\textcolor{blue}{0.8} &\textcolor{blue}{0.035} &0.74 &\textcolor{blue}{0.827} &\textcolor{blue}{0.1} &0.611 &\textcolor{blue}{0.711}\\
		(g) B+SRG+HTA+3DConv  &0.044 &0.747 &0.797 &0.038 &0.732 &0.817 &\textcolor{blue}{0.1} &0.613 &0.709\\
		\hline
		
		(h) B+SRG+HTA+LCFA	&\textcolor{red}{0.04}	&\textcolor{blue}{0.754} &\textcolor{red}{0.808}	&\textcolor{red}{0.03}	&\textcolor{blue}{0.753}	&\textcolor{red}{0.834} &\textcolor{red}{0.094} &0.622 &\textcolor{red}{0.718} \\
		\hline
\end{tabular}}
\caption{Performance of our ablation study experiments.} %
\label{ablation}
\end{table*}

\textbf{Quantitative Evaluation.}
We compare our model with our weakly supervised or unsupervised state-of-the-art salient object detection models on six benchmark datasets. As can be seen in Tab \ref{comparison_SOTA}, our model outperforms state-of-the-art weakly supervised methods by a large margin, albeit with much weaker supervision (scribble Anno. vs. point Anno.). Our model outperforms the previous best weakly supervised model \cite{wsvsod} by 2.1\% for $S_m$,  2.5\% for $F_\beta$, 0.9\% for MAE on average of six benchmark datasets. 
In particular, we outperform WSVSOD \cite{wsvsod} by 7.3\% for $F_\beta$, 1.6\% for $S_m$ and 1.7\% for MAE on the VOS \cite{vos_dataset} dataset.
What's more, our method even outperforms previous fully supervised methods SSAV \cite{ssav} on VOS dataset in all metrics.



\textbf{Qualitative Evaluation.}
As shown in Fig. \ref{vision_compare}, our model produces more complete and accurate saliency maps compared with other state-of-the-art weakly supervised methods and even exceeds recently fully supervised models.
Lines 1 and 2  illustrate our model can effectively distinguish video salient objects that are similar to the background. Lines 3 and 4 show that our method can overcome the interference of motion blur. The reason is that the attention-based method can effectively use global cues to reduce local and global interference. 
Line 5 demonstrates the ability of our model to detect small objects and objects in complex backgrounds. Line 6 illustrates that our method can effectively extract slender objects, which requires a strong ability to detect details. The reason is that our point supervision combined with flood filling can generate pseudo-labels with partial details, which complements the model with partial details.





\subsection{Ablation Study}
As shown in Tab. \ref{ablation}, we exhaustively analyze the proposed modules. "B" indicates the baseline model. "T2T-ViT\_t-14" is used as the backbone, and the features generated by the edge detection part are directly merged without using salient region guidance (SRG). ROFA and TA are both replaced by fully connected layers. Lines (a)-(e) are not equipped with multi-frame processing modules, and lines (f)-(h) have multi-frame processing modules.

		

\textbf{Salient Region Guidance.} Line (b) indicates that we add Salient Region guidance to the baseline model. Compared with line (a), (b) has better performance, the reason is that (b) explicitly filters out the edge of non-salient regions, reducing the interference of these useless edges on the results, which relieves the model from having to learn to filter out these distractions.

\textbf{Hybrid Token Attention.} Hybrid Token Attention (HTA) consists of RGB Optical Flow Attention (ROFA) and Token Attention (TA). Compared with (b), (c) and (d) improve the performance marg-inally, but when (c) and (d) are combined (i.e., Hybrid Token Attention (e)), the performance is significantly improved, even obtaining the best performance in some metrics. Compared to (b), (e) gains in all metrics, especially in the $F_\beta$ metric by 2.5\% on the DAVIS dataset and by 2.2\% on the SegTrack-V2 dataset. The improvement in consistency illustrates the effectiveness of our proposed Hybrid Token Attention. HTA can effectively fuse RGB image token and optical flow token in terms of short-term motion information.

\textbf{Long-term Cross-frame Attention.}
Line (h) indicates that on top of the (e), we add the LCFA module, which serves as our final model in this paper. ConvLSTM and 3D convolution are often used to process inter-frame information. Here, we replace the LCFA module with ConvLSTM and 3D convolution in lines (f) and (g), respectively. Compared with line (e), both lines (f) and (g) show performance gains. The reason is that they employ multiple frame information to eliminate ambiguity in the current frame. The model equipped with ConvLSTM (line (f)) is better than the model equipped with 3D convolution (line (g)). What's more, the model equipped with our proposed LCFA outperforms line (f) in almost all metrics, which shows the stronger ability of LCFA to exploit multi-frame information. Owing to the explicit correlation operation of the self-attention layers, LCFA can leverage the salient objects in the surrounding easy frames to find out the salient objects that are hard in the current frame, as shown in Fig. \ref{lcfa_visual}.

\begin{table}
\centering
\label{tab:freq}
	\scalebox{0.99}{
		\begin{tabular}{c|ccc| ccc}
			\hline
			\multirow{2}{*}{$\gamma$}    &\multicolumn{3}{c|}{DAVIS}   &\multicolumn{3}{c}{ViSal}\cr
			&MAE$\downarrow$  &$F_\beta \uparrow$ &$S_m \uparrow$  &MAE$\downarrow$  &$F_\beta \uparrow$ &$S_m \uparrow$\\
			\hline
			4  &0.058 &0.725   &0.784 &0.039 &0.830 &0.865\\
			5  &0.051 &0.733   &0.792 &0.034 &0.847 &0.870\\
			6  &0.04  &0.754   &0.808 &0.028 &0.858 &0.878\\
			\hline
	\end{tabular}}
	\caption{ The impact of $\gamma$.}
	\label{gama}
\end{table}

\textbf{Impact of the Hyperparameter $\gamma$.} 
We apply the adaptive flood filling method \cite{psod} to P-DAVIS and P-DAVSOD to obtain pseudo-labels. $\gamma$ is the adaptive radius of adaptive flood filling, and we verify the effect of the $\gamma$ on the resulting pseudo-labels. $\gamma$ = 5 is used in \cite{psod}, however, we obtain better performance when applying $\gamma$ = 6 (1/6 of the minimum value of length and width of the image) on P-DAVIS and P-DAVSOD as seen in Tab. \ref{gama}. The reason is that the images of video datasets are longer and wider, while salient objects occupy a smaller proportion of the images, so a small $\gamma$ can prevent the introduction of more noise.

\section{Conclusion}

In this paper, we propose a point-supervised video salient object detection model. A hybrid token attention module is proposed to fuse optical flow and image information from orthogonal directions to mine short-term motion information. Considering that in hard frames the model needs to employ multi-frame cues to infer the salient objects, we develop a Long-term Cross-Frame Attention module (LCFA) that leverages multi-frame tokens to help the current frame find the salient objects. What's more, we label two point-supervised datasets, P-DAVIS and P-DAVSOD. Extensive experiments illustrate our method outperforms existing state-of-the-art weakly-supervised methods.

\section*{Acknowledgments}
This work was supported by the National Natural Science Foundation of China (No.62072112), Scientific and Technological Innovation Action Plan of Shanghai Science and Technology Committee (No.20511103102), Fudan University-CIOMP Joint Fund (No. FC2019-005), Double First-class Construction Fund (No. XM03211178).




\bibliographystyle{ACM-Reference-Format}
\bibliography{sample-base}


\begin{thebibliography}{73}


\ifx \showCODEN    \undefined \def \showCODEN     #1{\unskip}     \fi
\ifx \showDOI      \undefined \def \showDOI       #1{#1}\fi
\ifx \showISBNx    \undefined \def \showISBNx     #1{\unskip}     \fi
\ifx \showISBNxiii \undefined \def \showISBNxiii  #1{\unskip}     \fi
\ifx \showISSN     \undefined \def \showISSN      #1{\unskip}     \fi
\ifx \showLCCN     \undefined \def \showLCCN      #1{\unskip}     \fi
\ifx \shownote     \undefined \def \shownote      #1{#1}          \fi
\ifx \showarticletitle \undefined \def \showarticletitle #1{#1}   \fi
\ifx \showURL      \undefined \def \showURL       {\relax}        \fi
\providecommand\bibfield[2]{#2}
\providecommand\bibinfo[2]{#2}
\providecommand\natexlab[1]{#1}
\providecommand\showeprint[2][]{arXiv:#2}

\bibitem[Ashish~Vaswani and Polosukhin(2017)]%
        {attent_is_all}
\bibfield{author}{\bibinfo{person}{Niki Parmar Jakob Uszkoreit Llion Jones
  Aidan N Gomez Łukasz~Kaiser Ashish~Vaswani, Noam~Shazeer} {and}
  \bibinfo{person}{Illia Polosukhin}.} \bibinfo{year}{2017}\natexlab{}.
\newblock \showarticletitle{Attention is all you need}. In
  \bibinfo{booktitle}{\emph{NIPS}}.
\newblock


\bibitem[Bearman et~al\mbox{.}(2016)]%
        {what_point}
\bibfield{author}{\bibinfo{person}{Amy Bearman}, \bibinfo{person}{Olga
  Russakovsky}, \bibinfo{person}{Vittorio Ferrari}, {and} \bibinfo{person}{Li
  Fei-Fei}.} \bibinfo{year}{2016}\natexlab{}.
\newblock \showarticletitle{What’s the point: Semantic segmentation with
  point supervision}. In \bibinfo{booktitle}{\emph{European conference on
  computer vision}}. Springer, \bibinfo{pages}{549--565}.
\newblock


\bibitem[Benenson et~al\mbox{.}(2019)]%
        {point_large}
\bibfield{author}{\bibinfo{person}{Rodrigo Benenson}, \bibinfo{person}{Stefan
  Popov}, {and} \bibinfo{person}{Vittorio Ferrari}.}
  \bibinfo{year}{2019}\natexlab{}.
\newblock \showarticletitle{Large-scale interactive object segmentation with
  human annotators}. In \bibinfo{booktitle}{\emph{Proceedings of the IEEE/CVF
  Conference on Computer Vision and Pattern Recognition}}.
  \bibinfo{pages}{11700--11709}.
\newblock


\bibitem[Carion et~al\mbox{.}(2020)]%
        {detr}
\bibfield{author}{\bibinfo{person}{Nicolas Carion}, \bibinfo{person}{Francisco
  Massa}, \bibinfo{person}{Gabriel Synnaeve}, \bibinfo{person}{Nicolas
  Usunier}, \bibinfo{person}{Alexander Kirillov}, {and} \bibinfo{person}{Sergey
  Zagoruyko}.} \bibinfo{year}{2020}\natexlab{}.
\newblock \showarticletitle{End-to-end object detection with transformers}. In
  \bibinfo{booktitle}{\emph{European Conference on Computer Vision}}. Springer,
  \bibinfo{pages}{213--229}.
\newblock


\bibitem[Cheng et~al\mbox{.}(2022)]%
        {implicit_cod}
\bibfield{author}{\bibinfo{person}{Xuelian Cheng}, \bibinfo{person}{Huan
  Xiong}, \bibinfo{person}{Deng-ping Fan}, \bibinfo{person}{Yiran Zhong},
  \bibinfo{person}{Mehrtash Harandi}, \bibinfo{person}{Tom Drummond}, {and}
  \bibinfo{person}{Zongyuan Ge}.} \bibinfo{year}{2022}\natexlab{}.
\newblock \showarticletitle{Implicit Motion Handling for Video Camouflaged
  Object Detection}.
\newblock \bibinfo{journal}{\emph{arXiv preprint arXiv:2203.07363}}
  (\bibinfo{year}{2022}).
\newblock


\bibitem[Cong et~al\mbox{.}(2018)]%
        {pr0.3}
\bibfield{author}{\bibinfo{person}{Runmin Cong}, \bibinfo{person}{Jianjun Lei},
  \bibinfo{person}{Huazhu Fu}, \bibinfo{person}{Ming-Ming Cheng},
  \bibinfo{person}{Weisi Lin}, {and} \bibinfo{person}{Qingming Huang}.}
  \bibinfo{year}{2018}\natexlab{}.
\newblock \showarticletitle{Review of visual saliency detection with
  comprehensive information}.
\newblock \bibinfo{journal}{\emph{IEEE Transactions on circuits and Systems for
  Video Technology}} (\bibinfo{year}{2018}), \bibinfo{pages}{2941--2959}.
\newblock


\bibitem[Deng et~al\mbox{.}(2009)]%
        {imagenet}
\bibfield{author}{\bibinfo{person}{Jia Deng}, \bibinfo{person}{Wei Dong},
  \bibinfo{person}{Richard Socher}, \bibinfo{person}{Li-Jia Li},
  \bibinfo{person}{Kai Li}, {and} \bibinfo{person}{Li Fei-Fei}.}
  \bibinfo{year}{2009}\natexlab{}.
\newblock \showarticletitle{Imagenet: A large-scale hierarchical image
  database}. In \bibinfo{booktitle}{\emph{2009 IEEE conference on computer
  vision and pattern recognition}}. Ieee, \bibinfo{pages}{248--255}.
\newblock


\bibitem[Dosovitskiy et~al\mbox{.}(2020)]%
        {vit}
\bibfield{author}{\bibinfo{person}{A. Dosovitskiy}, \bibinfo{person}{L. Beyer},
  \bibinfo{person}{A. Kolesnikov}, \bibinfo{person}{D. Weissenborn}, {and}
  \bibinfo{person}{N. Houlsby}.} \bibinfo{year}{2020}\natexlab{}.
\newblock \showarticletitle{An Image is Worth 16x16 Words: Transformers for
  Image Recognition at Scale}.
\newblock  (\bibinfo{year}{2020}).
\newblock


\bibitem[Fan et~al\mbox{.}(2017)]%
        {s-measure}
\bibfield{author}{\bibinfo{person}{Deng-Ping Fan}, \bibinfo{person}{Ming-Ming
  Cheng}, \bibinfo{person}{Yun Liu}, \bibinfo{person}{Tao Li}, {and}
  \bibinfo{person}{Ali Borji}.} \bibinfo{year}{2017}\natexlab{}.
\newblock \showarticletitle{Structure-measure: A new way to evaluate foreground
  maps}. In \bibinfo{booktitle}{\emph{Proceedings of the IEEE international
  conference on computer vision}}. \bibinfo{pages}{4548--4557}.
\newblock


\bibitem[Fan et~al\mbox{.}(2019a)]%
        {davsod}
\bibfield{author}{\bibinfo{person}{Deng-Ping Fan}, \bibinfo{person}{Wenguan
  Wang}, \bibinfo{person}{Ming-Ming Cheng}, {and} \bibinfo{person}{Jianbing
  Shen}.} \bibinfo{year}{2019}\natexlab{a}.
\newblock \showarticletitle{Shifting more attention to video salient object
  detection}. In \bibinfo{booktitle}{\emph{Proceedings of the IEEE/CVF
  Conference on Computer Vision and Pattern Recognition}}.
  \bibinfo{pages}{8554--8564}.
\newblock


\bibitem[Fan et~al\mbox{.}(2019b)]%
        {ssav}
\bibfield{author}{\bibinfo{person}{Deng-Ping Fan}, \bibinfo{person}{Wenguan
  Wang}, \bibinfo{person}{Ming-Ming Cheng}, {and} \bibinfo{person}{Jianbing
  Shen}.} \bibinfo{year}{2019}\natexlab{b}.
\newblock \showarticletitle{Shifting more attention to video salient object
  detection}. In \bibinfo{booktitle}{\emph{Proceedings of the IEEE/CVF
  Conference on Computer Vision and Pattern Recognition}}.
  \bibinfo{pages}{8554--8564}.
\newblock


\bibitem[Gao et~al\mbox{.}(2022)]%
        {psod}
\bibfield{author}{\bibinfo{person}{Shuyong Gao}, \bibinfo{person}{Wei Zhang},
  \bibinfo{person}{Yan Wang}, \bibinfo{person}{Qianyu Guo},
  \bibinfo{person}{Chenglong Zhang}, \bibinfo{person}{Yangji He}, {and}
  \bibinfo{person}{Wenqiang Zhang}.} \bibinfo{year}{2022}\natexlab{}.
\newblock \showarticletitle{Weakly-Supervised Salient Object Detection Using
  Point Supervison}. In \bibinfo{booktitle}{\emph{Proceedings of the AAAI
  Conference on Artificial Intelligence}}. \bibinfo{pages}{670--678}.
\newblock


\bibitem[Gu et~al\mbox{.}(2020)]%
        {vsod_pyramid}
\bibfield{author}{\bibinfo{person}{Yuchao Gu}, \bibinfo{person}{Lijuan Wang},
  \bibinfo{person}{Ziqin Wang}, \bibinfo{person}{Yun Liu},
  \bibinfo{person}{Ming-Ming Cheng}, {and} \bibinfo{person}{Shao-Ping Lu}.}
  \bibinfo{year}{2020}\natexlab{}.
\newblock \showarticletitle{Pyramid constrained self-attention network for fast
  video salient object detection}. In \bibinfo{booktitle}{\emph{Proceedings of
  the AAAI conference on artificial intelligence}}, Vol.~\bibinfo{volume}{34}.
  \bibinfo{pages}{10869--10876}.
\newblock


\bibitem[Hadizadeh and Baji{\'c}(2013)]%
        {videocompression1}
\bibfield{author}{\bibinfo{person}{Hadi Hadizadeh} {and}
  \bibinfo{person}{Ivan~V Baji{\'c}}.} \bibinfo{year}{2013}\natexlab{}.
\newblock \showarticletitle{Saliency-aware video compression}.
\newblock \bibinfo{journal}{\emph{IEEE Transactions on Image Processing}}
  \bibinfo{volume}{23}, \bibinfo{number}{1} (\bibinfo{year}{2013}),
  \bibinfo{pages}{19--33}.
\newblock


\bibitem[He et~al\mbox{.}(2016)]%
        {resnet}
\bibfield{author}{\bibinfo{person}{Kaiming He}, \bibinfo{person}{Xiangyu
  Zhang}, \bibinfo{person}{Shaoqing Ren}, {and} \bibinfo{person}{Jian Sun}.}
  \bibinfo{year}{2016}\natexlab{}.
\newblock \showarticletitle{Deep Residual Learning for Image Recognition}. In
  \bibinfo{booktitle}{\emph{Proceedings of the IEEE/CVF Conference on Computer
  Vision and Pattern Recognition}}. \bibinfo{pages}{770--778}.
\newblock


\bibitem[Hu et~al\mbox{.}(2018)]%
        {senet}
\bibfield{author}{\bibinfo{person}{Jie Hu}, \bibinfo{person}{Li Shen}, {and}
  \bibinfo{person}{Gang Sun}.} \bibinfo{year}{2018}\natexlab{}.
\newblock \showarticletitle{Squeeze-and-excitation networks}. In
  \bibinfo{booktitle}{\emph{Proceedings of the IEEE/CVF Conference on Computer
  Vision and Pattern Recognition}}. \bibinfo{pages}{7132--7141}.
\newblock


\bibitem[Ilg et~al\mbox{.}(2017)]%
        {flownet2}
\bibfield{author}{\bibinfo{person}{Eddy Ilg}, \bibinfo{person}{Nikolaus Mayer},
  \bibinfo{person}{Tonmoy Saikia}, \bibinfo{person}{Margret Keuper},
  \bibinfo{person}{Alexey Dosovitskiy}, {and} \bibinfo{person}{Thomas Brox}.}
  \bibinfo{year}{2017}\natexlab{}.
\newblock \showarticletitle{Flownet 2.0: Evolution of optical flow estimation
  with deep networks}. In \bibinfo{booktitle}{\emph{Proceedings of the IEEE
  conference on computer vision and pattern recognition}}.
  \bibinfo{pages}{2462--2470}.
\newblock


\bibitem[Itti(2004)]%
        {videocompression2}
\bibfield{author}{\bibinfo{person}{Laurent Itti}.}
  \bibinfo{year}{2004}\natexlab{}.
\newblock \showarticletitle{Automatic foveation for video compression using a
  neurobiological model of visual attention}.
\newblock \bibinfo{journal}{\emph{IEEE transactions on image processing}}
  \bibinfo{volume}{13}, \bibinfo{number}{10} (\bibinfo{year}{2004}),
  \bibinfo{pages}{1304--1318}.
\newblock


\bibitem[Ji et~al\mbox{.}(2021)]%
        {full_vos}
\bibfield{author}{\bibinfo{person}{Ge-Peng Ji}, \bibinfo{person}{Keren Fu},
  \bibinfo{person}{Zhe Wu}, \bibinfo{person}{Deng-Ping Fan},
  \bibinfo{person}{Jianbing Shen}, {and} \bibinfo{person}{Ling Shao}.}
  \bibinfo{year}{2021}\natexlab{}.
\newblock \showarticletitle{Full-duplex strategy for video object
  segmentation}. In \bibinfo{booktitle}{\emph{Proceedings of the IEEE/CVF
  International Conference on Computer Vision}}. \bibinfo{pages}{4922--4933}.
\newblock


\bibitem[Kr{\"a}henb{\"u}hl and Koltun(2011)]%
        {crf}
\bibfield{author}{\bibinfo{person}{Philipp Kr{\"a}henb{\"u}hl} {and}
  \bibinfo{person}{Vladlen Koltun}.} \bibinfo{year}{2011}\natexlab{}.
\newblock \showarticletitle{Efficient inference in fully connected crfs with
  gaussian edge potentials}. In \bibinfo{booktitle}{\emph{Advances in Neural
  Information Processing Systems}}. \bibinfo{pages}{109--117}.
\newblock


\bibitem[Lee et~al\mbox{.}(2021)]%
        {easyfram_vos}
\bibfield{author}{\bibinfo{person}{Youngjo Lee}, \bibinfo{person}{Hongje
  Seong}, {and} \bibinfo{person}{Euntai Kim}.} \bibinfo{year}{2021}\natexlab{}.
\newblock \showarticletitle{Iteratively Selecting an Easy Reference Frame Makes
  Unsupervised Video Object Segmentation Easier}.
\newblock \bibinfo{journal}{\emph{arXiv preprint arXiv:2112.12402}}
  (\bibinfo{year}{2021}).
\newblock


\bibitem[Li et~al\mbox{.}(2013)]%
        {segtrack}
\bibfield{author}{\bibinfo{person}{Fuxin Li}, \bibinfo{person}{Taeyoung Kim},
  \bibinfo{person}{Ahmad Humayun}, \bibinfo{person}{David Tsai}, {and}
  \bibinfo{person}{James~M Rehg}.} \bibinfo{year}{2013}\natexlab{}.
\newblock \showarticletitle{Video segmentation by tracking many figure-ground
  segments}. In \bibinfo{booktitle}{\emph{Proceedings of the IEEE international
  conference on computer vision}}. \bibinfo{pages}{2192--2199}.
\newblock


\bibitem[Li et~al\mbox{.}(2018c)]%
        {wsi}
\bibfield{author}{\bibinfo{person}{Guanbin Li}, \bibinfo{person}{Yuan Xie},
  {and} \bibinfo{person}{Liang Lin}.} \bibinfo{year}{2018}\natexlab{c}.
\newblock \showarticletitle{Weakly supervised salient object detection using
  image labels}. In \bibinfo{booktitle}{\emph{Thirty-second AAAI conference on
  artificial intelligence}}.
\newblock


\bibitem[Li et~al\mbox{.}(2018d)]%
        {fgrn}
\bibfield{author}{\bibinfo{person}{Guanbin Li}, \bibinfo{person}{Yuan Xie},
  \bibinfo{person}{Tianhao Wei}, \bibinfo{person}{Keze Wang}, {and}
  \bibinfo{person}{Liang Lin}.} \bibinfo{year}{2018}\natexlab{d}.
\newblock \showarticletitle{Flow guided recurrent neural encoder for video
  salient object detection}. In \bibinfo{booktitle}{\emph{Proceedings of the
  IEEE conference on computer vision and pattern recognition}}.
  \bibinfo{pages}{3243--3252}.
\newblock


\bibitem[Li et~al\mbox{.}(2019)]%
        {mga}
\bibfield{author}{\bibinfo{person}{Haofeng Li}, \bibinfo{person}{Guanqi Chen},
  \bibinfo{person}{Guanbin Li}, {and} \bibinfo{person}{Yizhou Yu}.}
  \bibinfo{year}{2019}\natexlab{}.
\newblock \showarticletitle{Motion guided attention for video salient object
  detection}. In \bibinfo{booktitle}{\emph{Proceedings of the IEEE/CVF
  International Conference on Computer Vision}}. \bibinfo{pages}{7274--7283}.
\newblock


\bibitem[Li et~al\mbox{.}(2017)]%
        {vos_dataset}
\bibfield{author}{\bibinfo{person}{Jia Li}, \bibinfo{person}{Changqun Xia},
  {and} \bibinfo{person}{Xiaowu Chen}.} \bibinfo{year}{2017}\natexlab{}.
\newblock \showarticletitle{A benchmark dataset and saliency-guided stacked
  autoencoders for video-based salient object detection}.
\newblock \bibinfo{journal}{\emph{IEEE Transactions on Image Processing}}
  \bibinfo{volume}{27}, \bibinfo{number}{1} (\bibinfo{year}{2017}),
  \bibinfo{pages}{349--364}.
\newblock


\bibitem[Li et~al\mbox{.}(2018b)]%
        {mbnm}
\bibfield{author}{\bibinfo{person}{Siyang Li}, \bibinfo{person}{Bryan Seybold},
  \bibinfo{person}{Alexey Vorobyov}, \bibinfo{person}{Xuejing Lei}, {and}
  \bibinfo{person}{C-C~Jay Kuo}.} \bibinfo{year}{2018}\natexlab{b}.
\newblock \showarticletitle{Unsupervised video object segmentation with
  motion-based bilateral networks}. In \bibinfo{booktitle}{\emph{Proceedings of
  the European conference on computer vision (ECCV)}}.
  \bibinfo{pages}{207--223}.
\newblock


\bibitem[Li et~al\mbox{.}(2020)]%
        {vsod_plug}
\bibfield{author}{\bibinfo{person}{Yunxiao Li}, \bibinfo{person}{Shuai Li},
  \bibinfo{person}{Chenglizhao Chen}, \bibinfo{person}{Aimin Hao}, {and}
  \bibinfo{person}{Hong Qin}.} \bibinfo{year}{2020}\natexlab{}.
\newblock \showarticletitle{A plug-and-play scheme to adapt image saliency deep
  model for video data}.
\newblock \bibinfo{journal}{\emph{IEEE Transactions on Circuits and Systems for
  Video Technology}} \bibinfo{volume}{31}, \bibinfo{number}{6}
  (\bibinfo{year}{2020}), \bibinfo{pages}{2315--2327}.
\newblock


\bibitem[Li et~al\mbox{.}(2018a)]%
        {point_latent}
\bibfield{author}{\bibinfo{person}{Zhuwen Li}, \bibinfo{person}{Qifeng Chen},
  {and} \bibinfo{person}{Vladlen Koltun}.} \bibinfo{year}{2018}\natexlab{a}.
\newblock \showarticletitle{Interactive image segmentation with latent
  diversity}. In \bibinfo{booktitle}{\emph{Proceedings of the IEEE Conference
  on Computer Vision and Pattern Recognition}}. \bibinfo{pages}{577--585}.
\newblock


\bibitem[Liew et~al\mbox{.}(2017)]%
        {point_regional}
\bibfield{author}{\bibinfo{person}{JunHao Liew}, \bibinfo{person}{Yunchao Wei},
  \bibinfo{person}{Wei Xiong}, \bibinfo{person}{Sim-Heng Ong}, {and}
  \bibinfo{person}{Jiashi Feng}.} \bibinfo{year}{2017}\natexlab{}.
\newblock \showarticletitle{Regional interactive image segmentation networks}.
  In \bibinfo{booktitle}{\emph{2017 IEEE international conference on computer
  vision (ICCV)}}. IEEE Computer Society, \bibinfo{pages}{2746--2754}.
\newblock


\bibitem[Liu et~al\mbox{.}(2019)]%
        {poolnet}
\bibfield{author}{\bibinfo{person}{Jiang-Jiang Liu}, \bibinfo{person}{Qibin
  Hou}, \bibinfo{person}{Ming-Ming Cheng}, \bibinfo{person}{Jiashi Feng}, {and}
  \bibinfo{person}{Jianmin Jiang}.} \bibinfo{year}{2019}\natexlab{}.
\newblock \showarticletitle{A simple pooling-based design for real-time salient
  object detection}. In \bibinfo{booktitle}{\emph{Proceedings of the IEEE/CVF
  Conference on Computer Vision and Pattern Recognition}}.
  \bibinfo{pages}{3917--3926}.
\newblock


\bibitem[Liu et~al\mbox{.}(2021c)]%
        {visual_saliency}
\bibfield{author}{\bibinfo{person}{Nian Liu}, \bibinfo{person}{Ni Zhang},
  \bibinfo{person}{Kaiyuan Wan}, \bibinfo{person}{Ling Shao}, {and}
  \bibinfo{person}{Junwei Han}.} \bibinfo{year}{2021}\natexlab{c}.
\newblock \showarticletitle{Visual saliency transformer}. In
  \bibinfo{booktitle}{\emph{Proceedings of the IEEE/CVF International
  Conference on Computer Vision}}. \bibinfo{pages}{4722--4732}.
\newblock


\bibitem[Liu et~al\mbox{.}(2021d)]%
        {visualsaliency}
\bibfield{author}{\bibinfo{person}{Nian Liu}, \bibinfo{person}{Ni Zhang},
  \bibinfo{person}{Kaiyuan Wan}, \bibinfo{person}{Ling Shao}, {and}
  \bibinfo{person}{Junwei Han}.} \bibinfo{year}{2021}\natexlab{d}.
\newblock \showarticletitle{Visual saliency transformer}. In
  \bibinfo{booktitle}{\emph{Proceedings of the IEEE/CVF International
  Conference on Computer Vision}}. \bibinfo{pages}{4722--4732}.
\newblock


\bibitem[Liu et~al\mbox{.}(2017)]%
        {edge_richer}
\bibfield{author}{\bibinfo{person}{Yun Liu}, \bibinfo{person}{Ming-Ming Cheng},
  \bibinfo{person}{Xiaowei Hu}, \bibinfo{person}{Kai Wang}, {and}
  \bibinfo{person}{Xiang Bai}.} \bibinfo{year}{2017}\natexlab{}.
\newblock \showarticletitle{Richer convolutional features for edge detection}.
  In \bibinfo{booktitle}{\emph{Proceedings of the IEEE conference on computer
  vision and pattern recognition}}. \bibinfo{pages}{3000--3009}.
\newblock


\bibitem[Liu et~al\mbox{.}(2021a)]%
        {swin}
\bibfield{author}{\bibinfo{person}{Z. Liu}, \bibinfo{person}{Y. Lin},
  \bibinfo{person}{Y. Cao}, \bibinfo{person}{H. Hu}, \bibinfo{person}{Y. Wei},
  \bibinfo{person}{Z. Zhang}, \bibinfo{person}{S. Lin}, {and}
  \bibinfo{person}{B. Guo}.} \bibinfo{year}{2021}\natexlab{a}.
\newblock \showarticletitle{Swin Transformer: Hierarchical Vision Transformer
  using Shifted Windows}.
\newblock  (\bibinfo{year}{2021}).
\newblock


\bibitem[Liu et~al\mbox{.}(2021b)]%
        {tritransnet}
\bibfield{author}{\bibinfo{person}{Zhengyi Liu}, \bibinfo{person}{Yuan Wang},
  \bibinfo{person}{Zhengzheng Tu}, \bibinfo{person}{Yun Xiao}, {and}
  \bibinfo{person}{Bin Tang}.} \bibinfo{year}{2021}\natexlab{b}.
\newblock \showarticletitle{Tritransnet: Rgb-d salient object detection with a
  triplet transformer embedding network}. In
  \bibinfo{booktitle}{\emph{Proceedings of the 29th ACM International
  Conference on Multimedia}}. \bibinfo{pages}{4481--4490}.
\newblock


\bibitem[Maninis et~al\mbox{.}(2018)]%
        {point_extreme}
\bibfield{author}{\bibinfo{person}{Kevis-Kokitsi Maninis},
  \bibinfo{person}{Sergi Caelles}, \bibinfo{person}{Jordi Pont-Tuset}, {and}
  \bibinfo{person}{Luc Van~Gool}.} \bibinfo{year}{2018}\natexlab{}.
\newblock \showarticletitle{Deep extreme cut: From extreme points to object
  segmentation}. In \bibinfo{booktitle}{\emph{Proceedings of the IEEE
  Conference on Computer Vision and Pattern Recognition}}.
  \bibinfo{pages}{616--625}.
\newblock


\bibitem[Obukhov et~al\mbox{.}(2019)]%
        {gate_crf}
\bibfield{author}{\bibinfo{person}{Anton Obukhov}, \bibinfo{person}{Stamatios
  Georgoulis}, \bibinfo{person}{Dengxin Dai}, {and} \bibinfo{person}{Luc
  Van~Gool}.} \bibinfo{year}{2019}\natexlab{}.
\newblock \showarticletitle{Gated CRF loss for weakly supervised semantic image
  segmentation}.
\newblock \bibinfo{journal}{\emph{arXiv preprint arXiv:1906.04651}}
  (\bibinfo{year}{2019}).
\newblock


\bibitem[Ochs et~al\mbox{.}(2013)]%
        {fbms}
\bibfield{author}{\bibinfo{person}{Peter Ochs}, \bibinfo{person}{Jitendra
  Malik}, {and} \bibinfo{person}{Thomas Brox}.}
  \bibinfo{year}{2013}\natexlab{}.
\newblock \showarticletitle{Segmentation of moving objects by long term video
  analysis}.
\newblock \bibinfo{journal}{\emph{IEEE transactions on pattern analysis and
  machine intelligence}} \bibinfo{volume}{36}, \bibinfo{number}{6}
  (\bibinfo{year}{2013}), \bibinfo{pages}{1187--1200}.
\newblock


\bibitem[Pan et~al\mbox{.}(2017)]%
        {videocaption}
\bibfield{author}{\bibinfo{person}{Yingwei Pan}, \bibinfo{person}{Ting Yao},
  \bibinfo{person}{Houqiang Li}, {and} \bibinfo{person}{Tao Mei}.}
  \bibinfo{year}{2017}\natexlab{}.
\newblock \showarticletitle{Video captioning with transferred semantic
  attributes}. In \bibinfo{booktitle}{\emph{Proceedings of the IEEE conference
  on computer vision and pattern recognition}}. \bibinfo{pages}{6504--6512}.
\newblock


\bibitem[Perazzi et~al\mbox{.}(2016)]%
        {davis16}
\bibfield{author}{\bibinfo{person}{Federico Perazzi}, \bibinfo{person}{Jordi
  Pont-Tuset}, \bibinfo{person}{Brian McWilliams}, \bibinfo{person}{Luc
  Van~Gool}, \bibinfo{person}{Markus Gross}, {and} \bibinfo{person}{Alexander
  Sorkine-Hornung}.} \bibinfo{year}{2016}\natexlab{}.
\newblock \showarticletitle{A benchmark dataset and evaluation methodology for
  video object segmentation}. In \bibinfo{booktitle}{\emph{Proceedings of the
  IEEE conference on computer vision and pattern recognition}}.
  \bibinfo{pages}{724--732}.
\newblock


\bibitem[Piao et~al\mbox{.}(2021)]%
        {mfnet}
\bibfield{author}{\bibinfo{person}{Yongri Piao}, \bibinfo{person}{Jian Wang},
  \bibinfo{person}{Miao Zhang}, {and} \bibinfo{person}{Huchuan Lu}.}
  \bibinfo{year}{2021}\natexlab{}.
\newblock \showarticletitle{MFNet: Multi-filter Directive Network for Weakly
  Supervised Salient Object Detection}. In
  \bibinfo{booktitle}{\emph{Proceedings of the IEEE/CVF International
  Conference on Computer Vision}}. \bibinfo{pages}{4136--4145}.
\newblock


\bibitem[Qian et~al\mbox{.}(2019)]%
        {metric_point}
\bibfield{author}{\bibinfo{person}{Rui Qian}, \bibinfo{person}{Yunchao Wei},
  \bibinfo{person}{Honghui Shi}, \bibinfo{person}{Jiachen Li},
  \bibinfo{person}{Jiaying Liu}, {and} \bibinfo{person}{Thomas Huang}.}
  \bibinfo{year}{2019}\natexlab{}.
\newblock \showarticletitle{Weakly supervised scene parsing with point-based
  distance metric learning}. In \bibinfo{booktitle}{\emph{Proceedings of the
  AAAI Conference on Artificial Intelligence}}, Vol.~\bibinfo{volume}{33}.
  \bibinfo{pages}{8843--8850}.
\newblock


\bibitem[Ren et~al\mbox{.}(2020)]%
        {tenet}
\bibfield{author}{\bibinfo{person}{Sucheng Ren}, \bibinfo{person}{Chu Han},
  \bibinfo{person}{Xin Yang}, \bibinfo{person}{Guoqiang Han}, {and}
  \bibinfo{person}{Shengfeng He}.} \bibinfo{year}{2020}\natexlab{}.
\newblock \showarticletitle{Tenet: Triple excitation network for video salient
  object detection}. In \bibinfo{booktitle}{\emph{European Conference on
  Computer Vision}}. Springer, \bibinfo{pages}{212--228}.
\newblock


\bibitem[Shi et~al\mbox{.}(2015)]%
        {convlstm}
\bibfield{author}{\bibinfo{person}{Xingjian Shi}, \bibinfo{person}{Zhourong
  Chen}, \bibinfo{person}{Hao Wang}, \bibinfo{person}{Dit-Yan Yeung},
  \bibinfo{person}{Wai-Kin Wong}, {and} \bibinfo{person}{Wang-chun Woo}.}
  \bibinfo{year}{2015}\natexlab{}.
\newblock \showarticletitle{Convolutional LSTM network: A machine learning
  approach for precipitation nowcasting}.
\newblock \bibinfo{journal}{\emph{Advances in neural information processing
  systems}}  \bibinfo{volume}{28} (\bibinfo{year}{2015}).
\newblock


\bibitem[Simonyan and Zisserman(2015)]%
        {vgg}
\bibfield{author}{\bibinfo{person}{Karen Simonyan} {and}
  \bibinfo{person}{Andrew Zisserman}.} \bibinfo{year}{2015}\natexlab{}.
\newblock \showarticletitle{Very deep convolutional networks for large-scale
  image recognition}. In \bibinfo{booktitle}{\emph{Proceedings of International
  Conference on Learning Representation}}.
\newblock


\bibitem[Song et~al\mbox{.}(2018)]%
        {pdb}
\bibfield{author}{\bibinfo{person}{Hongmei Song}, \bibinfo{person}{Wenguan
  Wang}, \bibinfo{person}{Sanyuan Zhao}, \bibinfo{person}{Jianbing Shen}, {and}
  \bibinfo{person}{Kin-Man Lam}.} \bibinfo{year}{2018}\natexlab{}.
\newblock \showarticletitle{Pyramid dilated deeper convlstm for video salient
  object detection}. In \bibinfo{booktitle}{\emph{Proceedings of the European
  conference on computer vision (ECCV)}}. \bibinfo{pages}{715--731}.
\newblock


\bibitem[Song et~al\mbox{.}(2021)]%
        {object_loca2}
\bibfield{author}{\bibinfo{person}{Qingyu Song}, \bibinfo{person}{Changan
  Wang}, \bibinfo{person}{Zhengkai Jiang}, \bibinfo{person}{Yabiao Wang},
  \bibinfo{person}{Ying Tai}, \bibinfo{person}{Chengjie Wang},
  \bibinfo{person}{Jilin Li}, \bibinfo{person}{Feiyue Huang}, {and}
  \bibinfo{person}{Yang Wu}.} \bibinfo{year}{2021}\natexlab{}.
\newblock \showarticletitle{Rethinking counting and localization in crowds: A
  purely point-based framework}. In \bibinfo{booktitle}{\emph{Proceedings of
  the IEEE/CVF International Conference on Computer Vision}}.
  \bibinfo{pages}{3365--3374}.
\newblock


\bibitem[Sun et~al\mbox{.}(2019)]%
        {pose_est_deep}
\bibfield{author}{\bibinfo{person}{Ke Sun}, \bibinfo{person}{Bin Xiao},
  \bibinfo{person}{Dong Liu}, {and} \bibinfo{person}{Jingdong Wang}.}
  \bibinfo{year}{2019}\natexlab{}.
\newblock \showarticletitle{Deep high-resolution representation learning for
  human pose estimation}. In \bibinfo{booktitle}{\emph{Proceedings of the
  IEEE/CVF Conference on Computer Vision and Pattern Recognition}}.
  \bibinfo{pages}{5693--5703}.
\newblock


\bibitem[Tang et~al\mbox{.}(2018a)]%
        {partial_ce}
\bibfield{author}{\bibinfo{person}{Meng Tang}, \bibinfo{person}{Abdelaziz
  Djelouah}, \bibinfo{person}{Federico Perazzi}, \bibinfo{person}{Yuri Boykov},
  {and} \bibinfo{person}{Christopher Schroers}.}
  \bibinfo{year}{2018}\natexlab{a}.
\newblock \showarticletitle{Normalized cut loss for weakly-supervised cnn
  segmentation}. In \bibinfo{booktitle}{\emph{Proceedings of the IEEE
  Conference on Computer Vision and Pattern Recognition}}.
  \bibinfo{pages}{1818--1827}.
\newblock


\bibitem[Tang et~al\mbox{.}(2018b)]%
        {vsod_weakly}
\bibfield{author}{\bibinfo{person}{Yi Tang}, \bibinfo{person}{Wenbin Zou},
  \bibinfo{person}{Zhi Jin}, \bibinfo{person}{Yuhuan Chen},
  \bibinfo{person}{Yang Hua}, {and} \bibinfo{person}{Xia Li}.}
  \bibinfo{year}{2018}\natexlab{b}.
\newblock \showarticletitle{Weakly supervised salient object detection with
  spatiotemporal cascade neural networks}.
\newblock \bibinfo{journal}{\emph{IEEE Transactions on Circuits and Systems for
  Video Technology}} \bibinfo{volume}{29}, \bibinfo{number}{7}
  (\bibinfo{year}{2018}), \bibinfo{pages}{1973--1984}.
\newblock


\bibitem[Wang et~al\mbox{.}(2021b)]%
        {panoptic_tr}
\bibfield{author}{\bibinfo{person}{Huiyu Wang}, \bibinfo{person}{Yukun Zhu},
  \bibinfo{person}{Hartwig Adam}, \bibinfo{person}{Alan Yuille}, {and}
  \bibinfo{person}{Liang-Chieh Chen}.} \bibinfo{year}{2021}\natexlab{b}.
\newblock \showarticletitle{Max-deeplab: End-to-end panoptic segmentation with
  mask transformers}. In \bibinfo{booktitle}{\emph{Proceedings of the IEEE/CVF
  Conference on Computer Vision and Pattern Recognition}}.
  \bibinfo{pages}{5463--5474}.
\newblock


\bibitem[Wang et~al\mbox{.}(2017)]%
        {duts}
\bibfield{author}{\bibinfo{person}{Lijun Wang}, \bibinfo{person}{Huchuan Lu},
  \bibinfo{person}{Yifan Wang}, \bibinfo{person}{Mengyang Feng},
  \bibinfo{person}{Dong Wang}, \bibinfo{person}{Baocai Yin}, {and}
  \bibinfo{person}{Xiang Ruan}.} \bibinfo{year}{2017}\natexlab{}.
\newblock \showarticletitle{Learning to detect salient objects with image-level
  supervision}. In \bibinfo{booktitle}{\emph{Proceedings of the IEEE/CVF
  Conference on Computer Vision and Pattern Recognition}}.
  \bibinfo{pages}{136--145}.
\newblock


\bibitem[Wang et~al\mbox{.}(2020)]%
        {crowd_count_nwpu}
\bibfield{author}{\bibinfo{person}{Qi Wang}, \bibinfo{person}{Junyu Gao},
  \bibinfo{person}{Wei Lin}, {and} \bibinfo{person}{Xuelong Li}.}
  \bibinfo{year}{2020}\natexlab{}.
\newblock \showarticletitle{NWPU-crowd: A large-scale benchmark for crowd
  counting and localization}.
\newblock \bibinfo{journal}{\emph{IEEE transactions on pattern analysis and
  machine intelligence}} \bibinfo{volume}{43}, \bibinfo{number}{6}
  (\bibinfo{year}{2020}), \bibinfo{pages}{2141--2149}.
\newblock


\bibitem[Wang et~al\mbox{.}(2015a)]%
        {trad_vsod1}
\bibfield{author}{\bibinfo{person}{Wenguan Wang}, \bibinfo{person}{Jianbing
  Shen}, {and} \bibinfo{person}{Fatih Porikli}.}
  \bibinfo{year}{2015}\natexlab{a}.
\newblock \showarticletitle{Saliency-aware geodesic video object segmentation}.
  In \bibinfo{booktitle}{\emph{Proceedings of the IEEE conference on computer
  vision and pattern recognition}}. \bibinfo{pages}{3395--3402}.
\newblock


\bibitem[Wang et~al\mbox{.}(2015b)]%
        {trad_vsod2}
\bibfield{author}{\bibinfo{person}{Wenguan Wang}, \bibinfo{person}{Jianbing
  Shen}, {and} \bibinfo{person}{Ling Shao}.} \bibinfo{year}{2015}\natexlab{b}.
\newblock \showarticletitle{Consistent video saliency using local gradient flow
  optimization and global refinement}.
\newblock \bibinfo{journal}{\emph{IEEE Transactions on Image Processing}}
  \bibinfo{volume}{24}, \bibinfo{number}{11} (\bibinfo{year}{2015}),
  \bibinfo{pages}{4185--4196}.
\newblock


\bibitem[Wang et~al\mbox{.}(2015c)]%
        {visal}
\bibfield{author}{\bibinfo{person}{Wenguan Wang}, \bibinfo{person}{Jianbing
  Shen}, {and} \bibinfo{person}{Ling Shao}.} \bibinfo{year}{2015}\natexlab{c}.
\newblock \showarticletitle{Consistent video saliency using local gradient flow
  optimization and global refinement}.
\newblock \bibinfo{journal}{\emph{IEEE Transactions on Image Processing}}
  \bibinfo{volume}{24}, \bibinfo{number}{11} (\bibinfo{year}{2015}),
  \bibinfo{pages}{4185--4196}.
\newblock


\bibitem[Wang et~al\mbox{.}(2021a)]%
        {pyramid_trs}
\bibfield{author}{\bibinfo{person}{Wenhai Wang}, \bibinfo{person}{Enze Xie},
  \bibinfo{person}{Xiang Li}, \bibinfo{person}{Deng-Ping Fan},
  \bibinfo{person}{Kaitao Song}, \bibinfo{person}{Ding Liang},
  \bibinfo{person}{Tong Lu}, \bibinfo{person}{Ping Luo}, {and}
  \bibinfo{person}{Ling Shao}.} \bibinfo{year}{2021}\natexlab{a}.
\newblock \showarticletitle{Pyramid vision transformer: A versatile backbone
  for dense prediction without convolutions}. In
  \bibinfo{booktitle}{\emph{Proceedings of the IEEE/CVF International
  Conference on Computer Vision}}. \bibinfo{pages}{568--578}.
\newblock


\bibitem[Wang et~al\mbox{.}(2018)]%
        {nonlocal}
\bibfield{author}{\bibinfo{person}{Xiaolong Wang}, \bibinfo{person}{Ross
  Girshick}, \bibinfo{person}{Abhinav Gupta}, {and} \bibinfo{person}{Kaiming
  He}.} \bibinfo{year}{2018}\natexlab{}.
\newblock \showarticletitle{Non-local neural networks}. In
  \bibinfo{booktitle}{\emph{Proceedings of the IEEE/CVF Conference on Computer
  Vision and Pattern Recognition}}. \bibinfo{pages}{7794--7803}.
\newblock


\bibitem[Yan et~al\mbox{.}(2019a)]%
        {vsod_semi}
\bibfield{author}{\bibinfo{person}{Pengxiang Yan}, \bibinfo{person}{Guanbin
  Li}, \bibinfo{person}{Yuan Xie}, \bibinfo{person}{Zhen Li},
  \bibinfo{person}{Chuan Wang}, \bibinfo{person}{Tianshui Chen}, {and}
  \bibinfo{person}{Liang Lin}.} \bibinfo{year}{2019}\natexlab{a}.
\newblock \showarticletitle{Semi-supervised video salient object detection
  using pseudo-labels}. In \bibinfo{booktitle}{\emph{Proceedings of the
  IEEE/CVF International Conference on Computer Vision}}.
  \bibinfo{pages}{7284--7293}.
\newblock


\bibitem[Yan et~al\mbox{.}(2019b)]%
        {rcrnet}
\bibfield{author}{\bibinfo{person}{Pengxiang Yan}, \bibinfo{person}{Guanbin
  Li}, \bibinfo{person}{Yuan Xie}, \bibinfo{person}{Zhen Li},
  \bibinfo{person}{Chuan Wang}, \bibinfo{person}{Tianshui Chen}, {and}
  \bibinfo{person}{Liang Lin}.} \bibinfo{year}{2019}\natexlab{b}.
\newblock \showarticletitle{Semi-supervised video salient object detection
  using pseudo-labels}. In \bibinfo{booktitle}{\emph{Proceedings of the
  IEEE/CVF International Conference on Computer Vision}}.
  \bibinfo{pages}{7284--7293}.
\newblock


\bibitem[Yu et~al\mbox{.}(2021)]%
        {scwssod}
\bibfield{author}{\bibinfo{person}{Siyue Yu}, \bibinfo{person}{Bingfeng Zhang},
  \bibinfo{person}{Jimin Xiao}, {and} \bibinfo{person}{Eng~Gee Lim}.}
  \bibinfo{year}{2021}\natexlab{}.
\newblock \showarticletitle{Structure-consistent weakly supervised salient
  object detection with local saliency coherence}. In
  \bibinfo{booktitle}{\emph{Proceedings of the AAAI Conference on Artificial
  Intelligence (AAAI)}}.
\newblock


\bibitem[Yu et~al\mbox{.}(2022)]%
        {object_loca1}
\bibfield{author}{\bibinfo{person}{Xuehui Yu}, \bibinfo{person}{Pengfei Chen},
  \bibinfo{person}{Di Wu}, \bibinfo{person}{Najmul Hassan},
  \bibinfo{person}{Guorong Li}, \bibinfo{person}{Junchi Yan},
  \bibinfo{person}{Humphrey Shi}, \bibinfo{person}{Qixiang Ye}, {and}
  \bibinfo{person}{Zhenjun Han}.} \bibinfo{year}{2022}\natexlab{}.
\newblock \showarticletitle{Object Localization under Single Coarse Point
  Supervision}.
\newblock \bibinfo{journal}{\emph{CVPR}} (\bibinfo{year}{2022}).
\newblock


\bibitem[Yuan et~al\mbox{.}(2021)]%
        {token_to_token}
\bibfield{author}{\bibinfo{person}{Li Yuan}, \bibinfo{person}{Yunpeng Chen},
  \bibinfo{person}{Tao Wang}, \bibinfo{person}{Weihao Yu},
  \bibinfo{person}{Yujun Shi}, \bibinfo{person}{Zi-Hang Jiang},
  \bibinfo{person}{Francis~EH Tay}, \bibinfo{person}{Jiashi Feng}, {and}
  \bibinfo{person}{Shuicheng Yan}.} \bibinfo{year}{2021}\natexlab{}.
\newblock \showarticletitle{Tokens-to-token vit: Training vision transformers
  from scratch on imagenet}. In \bibinfo{booktitle}{\emph{Proceedings of the
  IEEE/CVF International Conference on Computer Vision}}.
  \bibinfo{pages}{558--567}.
\newblock


\bibitem[Zeng et~al\mbox{.}(2019)]%
        {msw}
\bibfield{author}{\bibinfo{person}{Yu Zeng}, \bibinfo{person}{Yunzhi Zhuge},
  \bibinfo{person}{Huchuan Lu}, \bibinfo{person}{Lihe Zhang},
  \bibinfo{person}{Mingyang Qian}, {and} \bibinfo{person}{Yizhou Yu}.}
  \bibinfo{year}{2019}\natexlab{}.
\newblock \showarticletitle{Multi-source weak supervision for saliency
  detection}. In \bibinfo{booktitle}{\emph{Proceedings of the IEEE/CVF
  Conference on Computer Vision and Pattern Recognition}}.
  \bibinfo{pages}{6074--6083}.
\newblock


\bibitem[Zhang et~al\mbox{.}(2021a)]%
        {pose_est_ijcv}
\bibfield{author}{\bibinfo{person}{Jing Zhang}, \bibinfo{person}{Zhe Chen},
  {and} \bibinfo{person}{Dacheng Tao}.} \bibinfo{year}{2021}\natexlab{a}.
\newblock \showarticletitle{Towards high performance human keypoint detection}.
\newblock \bibinfo{journal}{\emph{International Journal of Computer Vision}}
  \bibinfo{volume}{129}, \bibinfo{number}{9} (\bibinfo{year}{2021}),
  \bibinfo{pages}{2639--2662}.
\newblock


\bibitem[Zhang et~al\mbox{.}(2020)]%
        {wssa}
\bibfield{author}{\bibinfo{person}{Jing Zhang}, \bibinfo{person}{Xin Yu},
  \bibinfo{person}{Aixuan Li}, \bibinfo{person}{Peipei Song},
  \bibinfo{person}{Bowen Liu}, {and} \bibinfo{person}{Yuchao Dai}.}
  \bibinfo{year}{2020}\natexlab{}.
\newblock \showarticletitle{Weakly-supervised salient object detection via
  scribble annotations}. In \bibinfo{booktitle}{\emph{Proceedings of the
  IEEE/CVF conference on computer vision and pattern recognition}}.
  \bibinfo{pages}{12546--12555}.
\newblock


\bibitem[Zhang et~al\mbox{.}(2021b)]%
        {dynamic}
\bibfield{author}{\bibinfo{person}{Miao Zhang}, \bibinfo{person}{Jie Liu},
  \bibinfo{person}{Yifei Wang}, \bibinfo{person}{Yongri Piao},
  \bibinfo{person}{Shunyu Yao}, \bibinfo{person}{Wei Ji},
  \bibinfo{person}{Jingjing Li}, \bibinfo{person}{Huchuan Lu}, {and}
  \bibinfo{person}{Zhongxuan Luo}.} \bibinfo{year}{2021}\natexlab{b}.
\newblock \showarticletitle{Dynamic context-sensitive filtering network for
  video salient object detection}. In \bibinfo{booktitle}{\emph{Proceedings of
  the IEEE/CVF International Conference on Computer Vision}}.
  \bibinfo{pages}{1553--1563}.
\newblock


\bibitem[Zhang et~al\mbox{.}(2021c)]%
        {dcfnet}
\bibfield{author}{\bibinfo{person}{Miao Zhang}, \bibinfo{person}{Jie Liu},
  \bibinfo{person}{Yifei Wang}, \bibinfo{person}{Yongri Piao},
  \bibinfo{person}{Shunyu Yao}, \bibinfo{person}{Wei Ji},
  \bibinfo{person}{Jingjing Li}, \bibinfo{person}{Huchuan Lu}, {and}
  \bibinfo{person}{Zhongxuan Luo}.} \bibinfo{year}{2021}\natexlab{c}.
\newblock \showarticletitle{Dynamic context-sensitive filtering network for
  video salient object detection}. In \bibinfo{booktitle}{\emph{Proceedings of
  the IEEE/CVF International Conference on Computer Vision}}.
  \bibinfo{pages}{1553--1563}.
\newblock


\bibitem[Zhao et~al\mbox{.}(2019)]%
        {egnet}
\bibfield{author}{\bibinfo{person}{Jiaxing Zhao}, \bibinfo{person}{Jiangjiang
  Liu}, \bibinfo{person}{Dengping Fan}, \bibinfo{person}{Yang Cao},
  \bibinfo{person}{Jufeng Yang}, {and} \bibinfo{person}{Mingming Cheng}.}
  \bibinfo{year}{2019}\natexlab{}.
\newblock \showarticletitle{EGNet: Edge Guidance Network for Salient Object
  Detection}. In \bibinfo{booktitle}{\emph{Proceedings of the IEEE/CVF
  International Conference on Computer Vision}}. \bibinfo{pages}{8779--8788}.
\newblock


\bibitem[Zhao et~al\mbox{.}(2021)]%
        {wsvsod}
\bibfield{author}{\bibinfo{person}{Wangbo Zhao}, \bibinfo{person}{Jing Zhang},
  \bibinfo{person}{Long Li}, \bibinfo{person}{Nick Barnes},
  \bibinfo{person}{Nian Liu}, {and} \bibinfo{person}{Junwei Han}.}
  \bibinfo{year}{2021}\natexlab{}.
\newblock \showarticletitle{Weakly Supervised Video Salient Object Detection}.
  In \bibinfo{booktitle}{\emph{Proceedings of the IEEE/CVF Conference on
  Computer Vision and Pattern Recognition}}. \bibinfo{pages}{16826--16835}.
\newblock


\bibitem[Zhou et~al\mbox{.}(2016)]%
        {cam}
\bibfield{author}{\bibinfo{person}{Bolei Zhou}, \bibinfo{person}{Aditya
  Khosla}, \bibinfo{person}{Agata Lapedriza}, \bibinfo{person}{Aude Oliva},
  {and} \bibinfo{person}{Antonio Torralba}.} \bibinfo{year}{2016}\natexlab{}.
\newblock \showarticletitle{Learning deep features for discriminative
  localization}. In \bibinfo{booktitle}{\emph{Proceedings of the IEEE
  conference on computer vision and pattern recognition}}.
  \bibinfo{pages}{2921--2929}.
\newblock


\bibitem[Zhu et~al\mbox{.}(2020)]%
        {deformable_detr}
\bibfield{author}{\bibinfo{person}{Xizhou Zhu}, \bibinfo{person}{Weijie Su},
  \bibinfo{person}{Lewei Lu}, \bibinfo{person}{Bin Li},
  \bibinfo{person}{Xiaogang Wang}, {and} \bibinfo{person}{Jifeng Dai}.}
  \bibinfo{year}{2020}\natexlab{}.
\newblock \showarticletitle{Deformable detr: Deformable transformers for
  end-to-end object detection}.
\newblock \bibinfo{journal}{\emph{arXiv preprint arXiv:2010.04159}}
  (\bibinfo{year}{2020}).
\newblock


\end{thebibliography}

\end{document}